%% file: main.tex
\documentclass[lettersize,journal]{IEEEtran}

\usepackage{comment}
\usepackage{amsmath,amssymb} 
\usepackage[accsupp]{axessibility}  
\usepackage[utf8]{inputenc} 
\usepackage[T1]{fontenc}    
\usepackage{url}            
\usepackage{booktabs}       
\usepackage[font=small,labelfont=bf]{caption}
\usepackage{subcaption}
\usepackage{amsfonts}       
\usepackage{nicefrac}       
\usepackage{microtype}      
\usepackage{times}
\usepackage{epsfig}
\usepackage[T1]{fontenc}
\usepackage{graphicx}
\usepackage{tabu}
\usepackage{textcomp}
\usepackage{gensymb}
\usepackage{wrapfig}
\usepackage{xcolor}
\usepackage{enumitem}
\usepackage[splitrule]{footmisc}
\usepackage{siunitx}
\usepackage{bm}
\usepackage{dirtytalk}
\usepackage{mathtools}
\usepackage{multirow}
\usepackage{wrapfig}
\usepackage{float}
\usepackage{tikz}

\usepackage[square,comma,numbers,sort&compress]{natbib}

\makeatletter 
\renewcommand\@biblabel[1]{#1.} 
\makeatother

\usepackage[pagebackref,breaklinks,colorlinks]{hyperref}
\hypersetup{
    linkcolor=cyan,
    urlcolor=cyan,
}
\usepackage[capitalize]{cleveref}
\crefname{section}{Sec.}{Secs.}
\Crefname{section}{Section}{Sections}
\Crefname{table}{Table}{Tables}
\crefname{table}{Tab.}{Tabs.}

\usepackage{orcidlink}

\let\emptyset\varnothing
\newcommand{\tobs}{\ensuremath{\bm{t}_\mathrm{obs}}}
\newcommand{\tfut}{\ensuremath{\bm{t}_\mathrm{fut}}}
\newcommand{\norm}[1]{\left\lVert #1 \right\rVert}
\newcommand{\ra}[1]{\renewcommand{\arraystretch}{#1}}

\begin{document}
\def\ECCVSubNumber{317}  

\title{Social Processes: Probabilistic Meta-learning for Adaptive Multiparty Interaction Forecasting}

%
\author{Augustinas Jučas \orcidlink{0009-0001-4412-2638} \and Chirag Raman \orcidlink{0000-0003-4894-4206}
\thanks{
Augustinas Jučas and Chirag Raman are with the Intelligent Systems, Delft University of Technology EEMCS, 225112 Delft, Zuid-Holland, Netherlands, 2628XE (e-mails: a.jucas@student.tudelft.nl, c.a.raman@tudelft.nl)
}
}
%
%

\maketitle


\begin{abstract}

Adaptively forecasting human behavior in social settings is an important step toward achieving Artificial General Intelligence. Most existing research in social forecasting has focused either on unfocused interactions, such as pedestrian trajectory prediction, or on monadic and dyadic behavior forecasting. In contrast, social psychology emphasizes the importance of group interactions for understanding complex social dynamics. This creates a gap that we address in this paper: forecasting social interactions at the group (conversation) level. Additionally, it is important for a forecasting model to be able to adapt to groups unseen at train time, as even the same individual behaves differently across different groups. This highlights the need for a forecasting model to \textit{explicitly} account for each group's unique dynamics. To achieve this, we adopt a \textit{meta-learning} approach to human behavior forecasting, treating every group as a separate \textit{meta-learning task}. As a result, our method conditions its predictions on the specific behaviors within the group, leading to generalization to unseen groups. Specifically, we introduce \textit{Social Process (SP)} models, which predict a \textit{distribution} over future multimodal cues \textit{jointly for all group members} based on their preceding low-level multimodal cues, while incorporating other past sequences of the same group's interactions. In this work we also analyze the generalization capabilities of SP models in both their outputs and latent spaces through the use of realistic synthetic datasets.

\end{abstract}


\section{Introduction}
\label{sec:introduction}

What does it take to develop Artificial General Intelligence (AGI) capable of interacting with humans in social environments? Humans, uniquely possessing general intelligence, depend heavily on their ability to anticipate the behavior of others to navigate complex interactions effectively \cite{kendonConductingInteractionPatterns1990, bohusManagingHumanRobotEngagement2014, bohusModelsMultipartyEngagement2009, ishiiPredictionNextUtteranceTiming2017, keitel2015use, garrod2015use, rochet2014take, Wlodarczak2016RespiratoryTC}. The ability to forecast social behavior even in novel situations is vital for tasks like turn-taking in conversations \cite{ishiiPredictionNextUtteranceTiming2017, keitel2015use, garrod2015use}, ensuring language comprehension and smooth conversational flow \cite{Gastaldon2024-gm}, coordinating mimicry episodes \cite{bilakhiaAudiovisualDetectionBehavioural2013}, and even regulating breathing patterns \cite{Wlodarczak2016RespiratoryTC}. 
Consequently, for machines to operate effectively in social settings and acquire adaptive social skills, they too must develop the capacity to predict the actions of those they interact with, especially in previously unseen social settings \cite{Klein2004-hi, Lemaignan2017-zb, Alami2006-ll}. This aligns with Endsley's model, which identifies the ability to project future states as the highest level of situational awareness \cite{Endsley1995-cq, Jiang2023-wj, Wei2020-hd}. Such dynamic predictive capabilities are already essential for applications ranging from pedestrian trajectory prediction \cite{rudenko2020human} to response generation \cite{Zhang2020-jq, OpenAI2023-ah} to assistive surgical robotics \cite{Zhou2018-fe}. Therefore it is evident that adaptive forecasting capability must be a crucial building block along the road to Artificial General Intelligence.
Most machine learning research on social interactions has concentrated on human behaviour modeling for single individuals \cite{He2024-sz, He2022-dk, Shafir2023-ud, Wang2023-dv} or dyadic interactions \cite{Ng2023-tw, Kucherenko2023-tg, Ng2022-ok, Palmero2021-bz, Ahuja2019-it}, with relatively little focus given to forecasting group dynamics. In contrast, social psychology has long emphasized the study of group interactions, which are essential for understanding complex social dynamics \cite{garrod2015use, keitel2015use}. Therefore, the limited exploration of specifically group forecasting in machine learning creates an important gap. To address this, our work focuses on forecasting social cues in conversational groups, bridging the gap between psychology and machine learning research. In particular, we focus on developing methods that can adapt to \textit{unseen} conversational groups, which is especially relevant in interaction forecasting where data is sparse and training separate models for different groups is not possible.


However, adapting to new conversational groups is difficult due to large variability in group interactions. For instance, in free-standing conversations, people actively sustain the social interactions by adapting to each other’s behaviors \citetext{\citealp[Chap.~1]{moore2013stacks}; \citealp[p.~237]{kendonConductingInteractionPatterns1990}}. 
As a result, even the behavior of the same individual can differ greatly across different groups, introducing significant complexity for forecasting models to adapt to. In fact, current state-of-the-art social forecasting models, which are typically trained in a standard supervised manner to predict behaviors for specific groups, struggle to generalize to new, unseen groups \cite{adeli2020socially, Tanke2023-zc, NEURIPS2021_2fd5d41e}. Therefore, we claim that the ability to adapt to new groups requires explicitly conditioning predictions on prior interactions within the same group. To achieve this, we propose a \textbf{meta-learning approach} that treats each group as a distinct meta-learning task. Such formulation allows to capture the social dynamics unique to each group without learning group-specific models and instead generalizing to unseen groups in a data-efficient manner.

There still remains a crucial question---what exactly should our models forecast? We are faced with effectively two options: a) forecasting high-level events in group interactions, such as group leaving, or b) forecasting low-level multimodal cues, such as human poses at every time step (see \autoref{fig:concept}). On the one hand, high-level forecasting has been an important topic of focus in social sciences, where researchers primarily employ a top-down workflow \cite{garrod2015use, keitel2015use}. On the other hand, low-level cue forecasting is significantly more data-efficient compared to high-level forecasting due to its self-supervised nature (see \autoref{fig:concept}). Additionally, low-level social cues retain their semantic meaning, allowing no information loss compared to higher-order event forecasting \cite{vinciarelliSocialSignalProcessing2009a}. In light of these observations, we choose to focus on low-level behavioral cues. Additionally, we observe that in group dynamics, the futures of interacting parties are all related \cite{kendonConductingInteractionPatterns1990}. Therefore, instead of treating every individual independently of others, a model should predict for all group members simultaneusly. Furthermore, since a single observed sequence can result in multiple socially valid futures, we allow for stochasticity in the predictions. With these constraints, we define a task called Social Cue Forecasting (SCF): predicting a \textit{distribution} over future multimodal cues \textit{jointly for all group members} from their same preceding multimodal cues.

As a result, we propose \textbf{Social Processes}: models, which aim to solve the SCF problem by taking the described meta-learning approach, conditioning the predicted distribution on previous interactions of that group. We believe that our framing of Social Cue Forecasting, as a \textit{few-shot} function estimation problem is especially suitable for conversation forecasting\textemdash a limited data regime which requires good uncertainty estimates as well as the ability to adapt to unseen groups. Therefore, we also perform a deep analysis into the generalization capabilities of the Social Process models: we employ a controlled semantic generalization analysis through the use of synthetic datasets. We also analyze the adaptation in the latent space, of both in-distribution and out-of-distribution samples.

Our contributions can be summarized as follows:
\begin{itemize}[itemsep=0em]
    \item We introduce and formalize the novel task of Social Cue Forecasting (SCF). 
    \item For SCF, we propose and evaluate the family of socially aware probabilistic Seq2Seq models we call Social Processes (SP). 
    \item We perform a deep analysis of generalization capabilities of the Social Process models both in the output and the latent space.
\end{itemize}

Note that this paper is an extension to "Social Processes: Self-Supervised Meta-Learning over Conversational Groups for Forecasting Nonverbal Social Cues", by Raman et al. \cite{Raman2023-cb}. In this paper, we extend the previous work with deep generalization analysis in synthetic settings as well as considerably stronger real-world experiments with larger and more expressive datasets.


\input{figures/concept}

\section{Related Work} \label{sec:relatedwork}
We review the current state of behavioral cue forecasting across three interaction scales: monanidc (single-person), dyadic (two-person), and multi-person interactions. Since body pose represents the primary non-verbal behavioral feature in humans, we also examine relevant work in human body pose estimation, which is vital for our contribution.

\textbf{Single Person Body Pose Modeling}. In recent years, significant advancements have been made in estimating human pose and motion from images and videos. Early works such as DeepPose \cite{Toshev2014-nh} and OpenPose \cite{Cao2021-rn} focused on estimating human skeleton keypoints directly from 2D image data. The introduction of the SMPL body model \cite{Loper2023-kn} standardized 3D human pose representation, later leading to enhanced variants such as SMPL-X \cite{Pavlakos2019-mi} and SMPL+H \cite{Romero2022-jm} that incorporate detailed modeling of hands, fingers, and facial expressions. This standardization lead to the development of numerous methods for fitting and refining SMPL parameters from video or keypoint data, including SMPLify \cite{Bogo2016-na}, SPIN \cite{Kolotouros2019-sg}, RoHM \cite{Zhang2024-bm}, EasyMocap \cite{easymocap}, and DeepMoCap \cite{Chatzitofis2019-km} among others. In our work, we employ EasyMocap to fit SMPL pose parameters to the CMU Panoptic Haggling dataset \cite{Joo_2017_TPAMI}. The field has also made significant advances in pose prior modeling, developing distributions over possible human poses that enable more robust pose estimation. These priors serve multiple additional purposes, including pose denoising, pose generation, and they can also work as regularization terms in optimization tasks, ultimately leading to more accurate and physically plausible pose regression \cite{Pratheepkumar2024-fq, Rempe2021-ic, Ci2023-px, Davydov2021-dh}.

\textbf{Monadic and Dyadic Behavioural Cue Forecasting}. The main interest in single-person behaviour forecasting is realistic \textit{motion} forecasting: predicting trajectories and poses, in various settings, such as a humans walking \cite{Martinez-Gonzalez2021-in}, gesturing \cite{ahujaStyleTransferCoSpeech2020}, dancing \cite{Li2021-ap, Guo2022-pq} and interacting with objects (reaching, grabbing) \cite{Martinez-Gonzalez2021-in}. In terms of dyadic interactions, the field of focus is wider, including speaking state transition forecasting \cite{kawahara2012prediction, Jiang2023-tg}, as well as body-related forecasting such as partner body pose alignment \cite{ahujaReactNotReact2019, palmero2022chalearn}, facial expressions \cite{palmero2022chalearn}, hand gestures \cite{ tuyen2022context, palmero2022chalearn}, gaze \cite{Muller2020-mz} and also higher-order characteristics such as anticipating intentions \cite{He2018-br} and human attention \cite{Cordel2019-no}.

\textbf{Multi-person Interactions}. Research on multi-person interaction forecasting encompasses both unfocused and focused interactions \cite{goffmanBehaviorPublicPlaces1966}. While substantial work exists on unfocused interactions, predicting pedestrian or vehicle trajectories \cite{helbingSocialForceModel1995, wasSocialDistancesModel2006, antoniniDiscreteChoiceModels2006, Treuille:2006:CC, robicquetLearningSocialEtiquette2016a, wangGaussianProcessDynamical2008, tayModellingSmoothPaths2007, pattersonIntentAwareProbabilisticTrajectory2019, alahiSocialLSTMHuman2016, zhangSRLSTMStateRefinement2019, guptaSocialGANSocially2018,hasanForecastingPeopleTrajectories2019,huangSTGATModelingSpatialTemporal2019, mohamedSocialSTGCNNSocialSpatioTemporal2020,zhaoTNTTargetdriveNTrajectory2020,gillesTHOMASTrajectoryHeatmap2022}, our primary interest lies in focused interactions, where participants actively coordinate their behaviors particularly in conversational settings. The social sciences have significantly influenced this field, leading to work predicting higher-level actions such as speaking turns \cite{garrod2015use, keitel2015use, rochet2014take, Wlodarczak2016RespiratoryTC}, interaction disengagement \cite{bohusManagingHumanRobotEngagement2014, vandoornRitualsLeavingPredictive2018}, and group dynamics such as splitting and merging \cite{wangGroupSplitMerge2020}. Some researchers have focused on predicting group size evolution \cite{mastrangeliRoundtableAbstractModel2010} or semantic social action labels \cite{airaleSocialInteractionGANMultipersonInteraction2021, sanghvi2019mgpi}. Recent years have seen emerging work on low-level non-verbal behavioral cue prediction, though this area remains underexplored. Earlier approaches of low-level cue forecasting predicted futures independently for each participant based on other participants' behaviors \cite{ahujaReactNotReact2019}. More recent methods forecast futures for all participants simultaneously, with Adeli et al. employing CNNs and GRUs for end-to-end pose prediction \cite{adeli2020socially}, Tanke et al. introducing diffusion models for future pose generation \cite{Tanke2023-zc}, and Wang et al. developing Transformer-based architectures \cite{NEURIPS2021_2fd5d41e} among others. While these works acknowledge that future states in focused group interactions are heavily group-dependent, to our knowledge, none have explored a meta-learning approach to enhance adaptability across different groups.

\medskip

\section{Social Cue Forecasting} \label{sec:task}

Self-supervision has proven effective for Large Language Models \cite{OpenAI2023-ah}, image \cite{Jaiswal2020-gq} and video data representations \cite{Schiappa2023-ko}. Importantly, the same self-supervised bottom-up approach is equally applicable to behavioral cues, as the semantic meaning conveyed in interactions (the so-called \textit{social signal} \cite{ambady2000toward}) is already embedded in low-level cues \cite{vinciarelliSocialSignalProcessing2009a}. Therefore, the representations of this high-level semantic meaning that we associate with actions and events (e.g. \textit{group leaving}) can be learned from the low-level dynamics in the cues and thus no information is lost when working with low-level cues. 


\subsection{Formalization and Distinction from Prior Task Formulations} \label{sec:task-reqs}
The objective of Social Cue Forecasting (SCF) is to predict future behavioral cues of \textit{all} people involved in a social encounter given an observed sequence of their behavioral features. Formally, let us denote a window of monotonically increasing observed timesteps as $\tobs~\coloneqq~[o1, o2, ..., oT]$, and an unobserved future time window as $\tfut~\coloneqq~[f1, f2, ..., fT]$, $f1>oT$. Note that
$\tfut$ and $\tobs$ can be of different lengths, and $\tfut$ does not need to immediately follow $\tobs$.
Given $n$ interacting participants, let us denote their social cues over $\tobs$ and $\tfut$ as
\begin{subequations}
\begin{gather}
    \bm{X} \coloneqq [\bm{b}^i_{t}; t \in \tobs]_{i=1}^n,\quad
    \bm{Y} \coloneqq [\bm{b}^i_{t}; t \in \tfut]_{i=1}^n. \tag{\theequation a, b}
\end{gather}
\end{subequations}
The vector $\bm{b}^i_{t}$ encapsulates the multimodal cues of interest from participant $i$ at time $t$. These can include head and body pose, speaking status, facial expressions, gestures, verbal content\textemdash any information streams that combine to transfer social meaning.

\textbf{Distribution over Futures}.
In its simplest form, given an $\bm{X}$, the objective of SCF is to learn a single function $f$ such that $\bm{Y} = f(\bm{X})$. However, an inherent challenge in forecasting behavior is that
an observed sequence of interaction does not have a deterministic future and can result in multiple socially valid ones\textemdash
a window of overlapping speech between people may and may not
result in a change of speaker \cite{heldnerPausesGapsOverlaps2010, duncanSignalsRulesTaking1972}, a change in head orientation may continue into a sweeping
glance across the room or a darting glance stopping at a
recipient of interest \cite{mooreNonverbalCourtshipPatterns1985}.
In some cases, certain observed behaviors\textemdash intonation and gaze cues
\cite{keitel2015use, kalmaGazingTriadsPowerful1992} or synchronization in speaker-listener speech
\cite{levinsonTimingTurntakingIts2015} for turn-taking\textemdash
may make some outcomes more likely than others. Given that there are both
supporting and challenging arguments for how these observations influence subsequent behaviors
\citetext{\citealp[p.~5]{levinsonTimingTurntakingIts2015}; \citealp[p.~22]{kalmaGazingTriadsPowerful1992}},
it would be beneficial if a data-driven model expresses a measure of uncertainty in its forecasts.
We do this by modeling the distribution over possible futures $p(\bm{Y}|\bm{X})$, rather than a single future $\bm{Y}$ for a given $\bm{X}$, the latter being the case for previous formulations for cues \cite{jooSocialArtificialIntelligence2019, ahujaReactNotReact2019, yaoMultipleGranularityGroup2018} and actions \cite{sanghvi2019mgpi, airaleSocialInteractionGANMultipersonInteraction2021}.

\textbf{Joint Modeling of Future Uncertainty}.
A defining characteristic of focused interactions is that the 
participants sustain the shared interaction through explicit, cooperative coordination of behavior \citetext{\citealp[p.~220]{kendonConductingInteractionPatterns1990}}\textemdash the futures of interacting individuals are not independent given an observed window of group behavior. It is therefore essential to capture uncertainty in forecasts at the \textit{global} level\textemdash jointly forecasting one future for all participants at
a time, rather than at a \textit{local} output level\textemdash one future for each individual independent of the remaining participants' futures. In contrast, applying the prior formulations \cite{ahujaReactNotReact2019, jooSocialArtificialIntelligence2019, palmero2022chalearn} requires the training of separate models treating each individual as a target (for the same group input) and then forecasting an independent future one at a time. Meanwhile, other prior pose forecasting works \cite{chaoForecastingHumanDynamics2017,fragkiadakiRecurrentNetworkModels2015, walkerPoseKnowsVideo2017, habibieRecurrentVariationalAutoencoder2017, pavlloQuaterNetQuaternionbasedRecurrent2018} have been in non-social settings and do not need to model such behavioral interdependence.

\textbf{Non-Contiguous Observed and Future Windows}. Domain experts are often interested in settings where $\tobs$ and $\tfut$ are offset by an arbitrary delay, such as forecasting a time lagged synchrony \cite{delahercheInterpersonalSynchronySurvey2012} or mimicry \cite{bilakhiaAudiovisualDetectionBehavioural2013} episode, or upcoming disengagement \cite{bohusManagingHumanRobotEngagement2014, vandoornRitualsLeavingPredictive2018}. We therefore allow for non-contiguous $\tobs$ and $\tfut$. Operationalizing prior formulations that predict one step into the future \cite{yaoMultipleGranularityGroup2018, sanghvi2019mgpi, jooSocialArtificialIntelligence2019, airaleSocialInteractionGANMultipersonInteraction2021} would entail a sliding window of autoregressive predictions over the offset between $\tobs$ and $\tfut$ (from $oT$ to $f1$), with errors cascading even before decoding is performed over the window of interest $\tfut$.

\medskip
\noindent Our task formalization of SCF can be viewed as a social science-grounded generalization of prior computational formulations, and therefore suitable for a wider range of cross-disciplinary tasks, both computational and analytical.

\section{Method Preliminaries} \label{sec:background}

\textbf{Meta-learning}. A standard supervised setting can be seen as learning a predictor $f(x)$ from a static labeled dataset $C \coloneqq (\bm{X}_C, \bm{Y}_C) \coloneqq \{(\bm{x}^i, \bm{y}^i)\}_{i \in \{1, \ldots, |C|\}}$. However, in a meta-learning setting, every task instance includes a small dataset $C$ and the goal is, given any task, to output a predictor based on that dataset. Therefore, training a meta-learning model requires learning to dynamically map from a dataset to a predictor. More formally the goal is to learn a mapping $C \mapsto f(\cdot,C)$.
In meta-learning literature, a \textit{task} refers to each dataset in a collection $\{\mathcal{T}_m\}_{m=1}^{N_{\mathrm{tasks}}}$ of related datasets \cite{hospedalesMetaLearningNeuralNetworks2020}. 
Every training iteration includes a single task $\mathcal{T}$, which is split into two subsets $(C, D)$, which are typically called \textit{context} and \textit{target} respectively.
A meta-learner aims to fit the subset of target points $D$ given the subset of context observations $C$. At meta-test time, the resulting predictor $f(\bm{x},C)$ is adapted to make predictions for target points on  an unseen task by conditioning on a new context set $C$ unseen during meta-training.


\textbf{Neural Processes (NPs)}. Sharing the same core motivations, NPs \cite{garneloNeuralProcesses2018} can be viewed as a family of latent variable models that extend the idea of meta-learning to situations where modeling epistemic uncertainty in the predictions $f(\bm{x},C)$ is desirable. NPs do this by meta-learning a map from datasets to stochastic processes, estimating a distribution over the predictions $p(\bm{Y}|\bm{X},C)$. To capture this distribution, NPs model the conditional latent distribution $p(\bm{z}|C)$ from which a task representation $\bm{z} \in \mathbb{R}^d$ is sampled. This introduces stochasticity, constituting what is called the model's \textit{latent path}. The context can also be directly incorporated through a \textit{deterministic path}, via a representation $\bm{r}_C \in \mathbb{R}^d$ aggregated over $C$. An observation model $p(\bm{y}^i|\bm{x}^i,\bm{r}_C,\bm{z})$ then fits the target observations in $D$.
The generative process for the NP is written as
\begin{equation}\label{eq:np}
    \begin{split}
    p(\bm{Y}|\bm{X},C) & \coloneqq \int p(\bm{Y}|\bm{X},C,\bm{z})p(\bm{z}|C)d\bm{z}  \\
    & =  \int p(\bm{Y}|\bm{X},\bm{r}_C,\bm{z})q(\bm{z}|\bm{s}_C)d\bm{z},
    \end{split}
\end{equation}
where  $p(\bm{Y}|\bm{X},\bm{r}_C,\bm{z}) \coloneqq \prod_{i \in \{1, ..., |D|\}} p(\bm{y}^i|\bm{x}^i,\bm{r}_C,\bm{z})$. The latent $\bm{z}$ is modeled by a factorized Gaussian parameterized by $\bm{s}_C \coloneqq f_s(C)$, with $f_s$ being a deterministic function invariant to order permutation over $C$. When the conditioning on context is removed $(C = \emptyset$), we have $q(\bm{z}|\bm{s}_\emptyset) \coloneqq p(\bm{z})$, the zero-information prior on $\bm{z}$. The deterministic path uses a function $f_r$ similar to $f_s$, so that $\bm{r}_C \coloneqq f_r(C)$. In practice this is implemented as $\bm{r}_C = \frac{1}{|C|} \sum_{i \in \{1, ..., |C|\}}\mathrm{MLP}(\bm{x}_i, \bm{y}_i)$.
The observation model is referred to as the \textit{decoder}, and
$q, f_r, f_s$ comprise the \textit{encoders}. The parameters of the NP are learned for random subsets $C$ and $D$ for a task by maximizing the evidence lower bound (ELBO)
\begin{equation}
    \begin{split}
     \log p(\bm{Y}|\bm{X},C) \ \geq \ \ &\mathbb{E}_{q(\bm{z}|\bm{s}_D)}[\log p(\bm{Y}|\bm{X},C,\bm{z})] \ - \\
     &\mathbb{KL}(q(\bm{z}|\bm{s}_D)\ || \ q(\bm{z}|\bm{s}_C)). \label{eq:elbo}
    \end{split}
\end{equation}

\section{Social Processes: Methodology} \label{sec:methodology}
Our core idea for adapting predictions to a group’s unique behavioral dynamics is to condition forecasts on a context set $C$ of the same group's observed-future sequence pairs.
By \textit{learning to learn}, i.e., \textit{meta-learn} from a context set, our model can generalize to unseen groups at evaluation by conditioning on an unseen context set of the test group's behavior sequences. In practice, a social robot might, for instance, observe such an evaluation context set before approaching a new group.

We set up by splitting the interaction into pairs of observed and future sequences, writing the context
as $C \coloneqq (\bm{X}_C, \bm{Y}_C) \coloneqq (\bm{X}_j, \bm{Y}_k)_{(j, k) \in [N_C] \times [N_C]}$, where every $\bm{X}_j$ occurs before the corresponding $\bm{Y}_k$. Since we allow for non-contiguous $\tobs$ and $\tfut$, the $j$th $\tobs$ can have multiple associated $\tfut$ windows for prediction, up to a maximum offset. Denoting the set of target window pairs as $D \coloneqq (\bm{X}, \bm{Y}) \coloneqq (\bm{X}_j, \bm{Y}_k)_{(j, k) \in [N_D]\times[N_D]}$, our goal is to model the distribution $p(\bm{Y}|\bm{X},C)$. Note that when conditioning on context is removed ($C = \emptyset$), we simply revert to the non-meta-learning formulation $p(\bm{Y}|\bm{X})$.

\begin{figure*}[!t]
\makebox[\textwidth][c]{\includegraphics[width=0.75\textwidth]{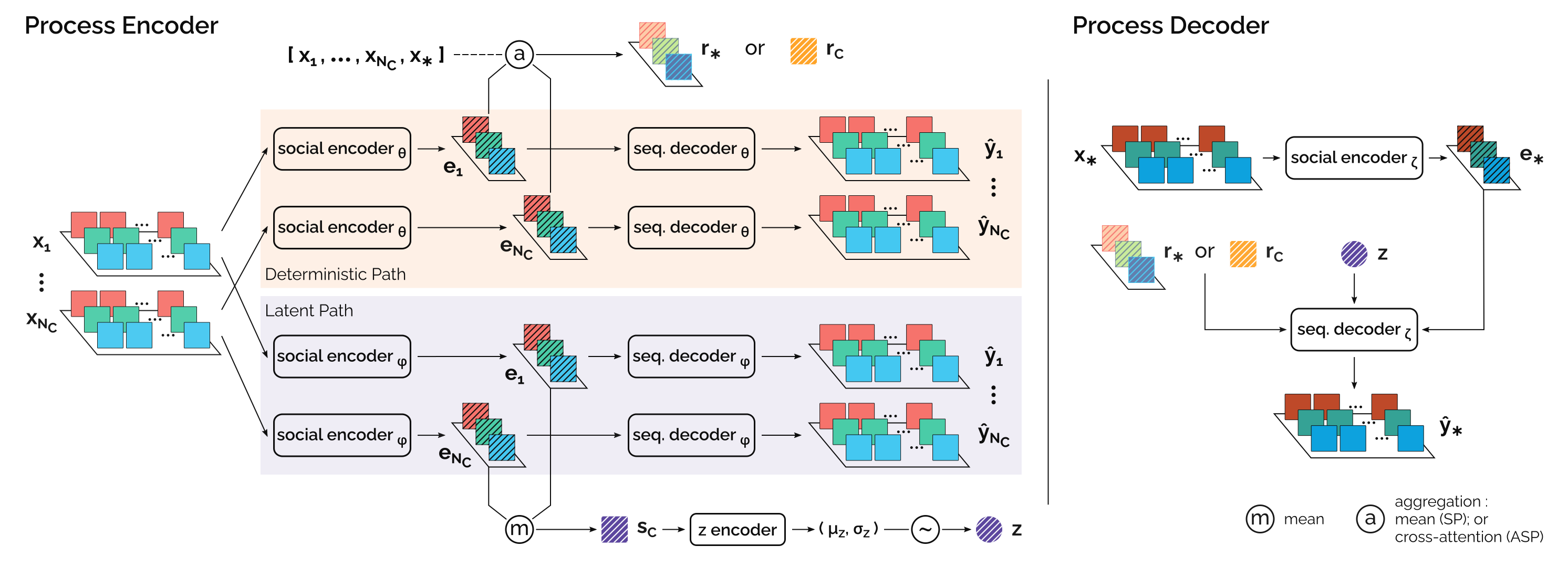}}
\caption{Architecture of the SP and ASP family.}
\label{fig:architecture}
\end{figure*}

The generative process for our Social Process (SP) model follows \autoref{eq:np}, which we extend to social forecasting in two ways. We embed an observed sequence $\bm{x}^i$ for participant $\mathrm{p}_i$ into a condensed encoding $\bm{e^i} \in \mathbb{R}^{d}$ that is then decoded into the future sequence using a Seq2Seq architecture \cite{sutskeverSequenceSequenceLearning2014, choLearningPhraseRepresentations2014}. Crucially, the sequence decoder only accesses $\bm{x}^i$ through $\bm{e}^i$. So after training, $\bm{e}^i$ must encode the \textit{temporal} information that $\bm{x}^i$ contains about the future. Further, social behavior is interdependent. We model $\bm{e}^i$ as a function of both, $\mathrm{p}_i$'s own behavior as well as that of partners $\mathrm{p}_{j, j\neq i}$ from $\mathrm{p}_i$'s perspective. This captures the \textit{spatial} influence partners have on the participant over $\tobs$. Using notation we established in \Cref{sec:task}, we define the observation model for $\mathrm{p}_i$ as
\begin{equation}
    \begin{split}
        p(\bm{y}^i|\bm{x}^i,C,\bm{z})
        &\coloneqq p(\bm{b}_{f1}^i,\ldots,\bm{b}_{fT}^i | \bm{b}_{o1}^i,\ldots,\bm{b}_{oT}^i,C,\bm{z})  \\
        &= p(\bm{b}_{f1}^i,\ldots,\bm{b}_{fT}^i | \bm{e}^i,\bm{r}_C,\bm{z}).\label{eq:spdef}
    \end{split}
\end{equation}
If decoding is carried out in an auto-regressive manner, the right hand side of \autoref{eq:spdef} simplifies to $\prod_{t=f1}^{fT} p(\bm{b}_{t}^i|\bm{b}_{t-1}^i, \ldots, \bm{b}_{f1}^i, \bm{e}^i,\bm{r}_C,\bm{z})$. Following the standard NP setting, we implement the observation model as a set of Gaussian distributions factorized over time and feature dimensions. We also incorporate the cross-attention mechanism from the Attentive Neural Process (ANP) \cite{kimAttentiveNeuralProcesses2019} to define the variant Attentive Social Process (ASP). Following \autoref{eq:spdef} and the definition of the ANP, the corresponding observation model of the ASP for a single participant is defined as
\begin{equation}
    p(\bm{y}^i|\bm{x}^i,C,\bm{z}) = p(\bm{b}_{f1}^i,\ldots,\bm{b}_{fT}^i|\bm{e}^i,r^*(C, \bm{x}^i),\bm{z}). \label{eq:aspdef}
\end{equation}
Here each target query sequence $\bm{x}_*^i$ attends to the context sequences $\bm{X}_C$ to produce a query-specific representation $\bm{r}_* \coloneqq r^*(C, \bm{x}_*^i) \in \mathbb{R}^{d}$.

The model architectures are illustrated in \autoref{fig:architecture}. Note that our modeling assumption is that the underlying stochastic process generating social behaviors does not evolve over time. That is, the individual factors determining how participants coordinate behaviors\textemdash age, cultural background, personality variables \citetext{\citealp[Chap.~1]{moore2013stacks}; \citealp[p.~237]{kendonConductingInteractionPatterns1990}}\textemdash are likely to remain the same over a single interaction. This is in contrast to the line of work that deals with \textit{meta-transfer learning}, where the stochastic process itself changes over time \cite{singhSequentialNeuralProcesses2019, yoonRobustifyingSequentialNeural2020, williRecurrentNeuralProcesses2019, kumarSpatiotemporalModelingUsing}; this entails modeling a different $\bm{z}$ distribution for every timestep.

\textbf{Encoding Partner Behavior}. To encode partners' influence on an individual's future, we use a pair of sequence encoders: one to encode the temporal dynamics of participant $\mathrm{p}^i$'s features, $\bm{e}_\mathrm{self}^i = f_\mathrm{self}(\bm{x}^i)$, and another to encode the dynamics of a transformed representation of the features of $\mathrm{p}^i$'s partners, $\bm{e}_\mathrm{partner}^i = f_\mathrm{partner}(\psi(\bm{x}^{j, (j \neq i)}))$. Using a separate network to encode partner behavior enables sampling an individual's and partners' features at different sampling rates.

\begin{figure}[!t]
\makebox[\columnwidth][c]{\includegraphics[width=\columnwidth]{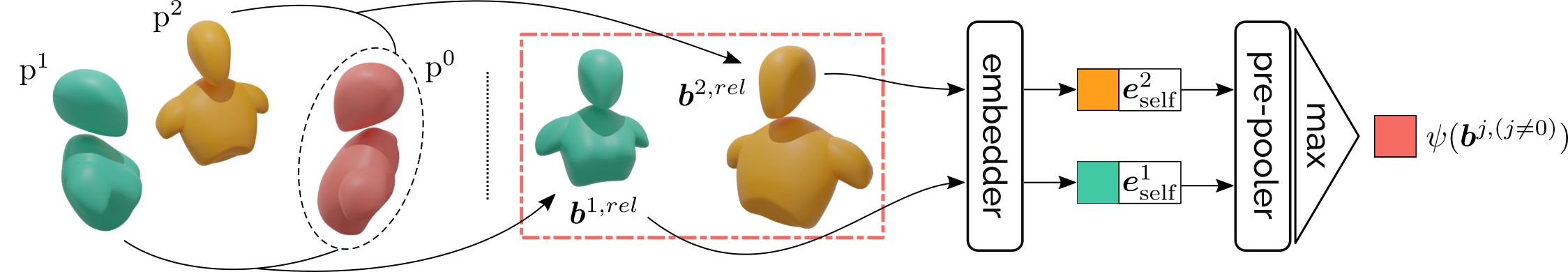}}
\caption{Encoding partner behavior for conversation participant $\mathrm{p}^0$ for a single timestep. To model the influence partners  $\mathrm{p}^1$ and $\mathrm{p}^2$ have on the behavior of $\mathrm{p}^0$, we transform the partner features to capture the interaction from $\mathrm{p}^0$'s perspective, and learn a representation of these features invariant to group size and partner-order permutation using the symmetric $\mathrm{max}$ function.}
\label{fig:pooling}
\end{figure}

How do we model $\psi(\bm{x}^{j, (j \neq i)})$? We want the partners' representation to possess two properties: \textit{permutation invariance}\textemdash changing the order of the partners should not affect the representation, and \textit{group-size independence}\textemdash we want to compactly represent all partners independent of the group size. Intuitively, to model partner influence on $\mathrm{p}^i$, we wish to \textit{capture a view of the partners' behavior as $\mathrm{p}^i$ perceives it}. Figure~\ref{fig:pooling} illustrates the underlying intuition. We do this by computing pooled embeddings of relative behavioral features, extending \citet{guptaSocialGANSocially2018}'s approach for pedestrian positions to conversation behavior. Note that our partner-encoding approach is in contrast to that of \citet{tanMultimodalJointHead2021}, which is order and group-size dependent, and \citet{yaoMultipleGranularityGroup2018}, who do not transform the partner features to an individual's perspective.

Since the most commonly considered cues in literature are pose (orientation and location) and binary speaking status \cite{alameda-pinedaAnalyzingFreestandingConversational2015, zhangSocialInvolvementMingling2018, tanMultimodalJointHead2021}, we specify how we transform them. For a single timestep, we denote these cues for $\mathrm{p}^i$ as $\bm{b}^i = [\bm{q}^i;\bm{l}^i;s^i]$, and for $\mathrm{p}^j$ as $\bm{b}^j = [\bm{q}^j;\bm{l}^j;s^j]$. We compute the relative partner features $\bm{b}^{j, rel} = [\bm{q}^{rel};\bm{l}^{rel};s^{rel}]$ by transforming $\bm{b}^j$ to a frame of reference defined by $\bm{b}^i$:
\begin{subequations}
\begin{gather}
    \bm{q}^{rel} = \bm{q}^i * (\bm{q}^j)^{-1}, \quad
    \bm{l}^{rel} = \bm{l}^j - \bm{l}^i,  \quad
    s^{rel} = s^j - s^i. \tag{\theequation a-c}
\end{gather}
\end{subequations}
Note that we use unit quaternions (denoted $\bm{q}$) for representing orientation due to their various benefits over other representations of rotation \citep[Sec.~3.2]{kendallGeometricLossFunctions2017}. The operator $*$ denotes the Hamilton product of the quaternions. These transformed features $\bm{b}^{j, rel}$ for each $\mathrm{p}^j$ are then encoded using an \textit{embedder} MLP. The outputs are concatenated with their corresponding $\bm{e}_\mathrm{self}^j$ and processed by a \textit{pre-pooler} MLP. Assuming  $d_\mathrm{in}$ and $d_\mathrm{out}$ pre-pooler input and output dims and $J$ partners, we stack the $J$ inputs to obtain $(J, d_\mathrm{in})$ tensors. The $(J, d_\mathrm{out})$-dim output is element-wise max-pooled over the $J$ dim, resulting in the $d_\mathrm{out}$-dim vector $\psi(\bm{b}^{j, (j \neq i)})$ for any value of $J$, per timestep. We capture the temporal dynamics in this pooled representation over $\tobs$ using $f_\mathrm{partner}$. Finally, we combine $\bm{e}_\mathrm{self}^i$ and $\bm{e}_\mathrm{partner}^i$ for $\mathrm{p}^i$ through a linear projection (defined by a weight matrix $W$) to obtain the individual's embedding $\bm{e}_\mathrm{ind}^i = W\cdot [\bm{e}_\mathrm{self}^i;\bm{e}_\mathrm{partner}^i]$. Our intuition is that with information about both $\mathrm{p}^i$ themselves, and of $\mathrm{p}^i$'s partners from $\mathrm{p}^i$'s point-of-view,  $\bm{e}_\mathrm{ind}^i$ now contains the information required to predict $\mathrm{p}^i$'s future behavior.

\textbf{Encoding Future Window Offset}. Since we allow for non-contiguous windows, a single $\tobs$ might be associated to multiple $\tfut$ windows at different offsets. Decoding the same $\bm{e}_\mathrm{ind}^i$ into multiple sequences (for different $\tfut$) in the absence of any timing information might cause an averaging effect in either the decoder or the information encoded in $\bm{e}_\mathrm{ind}^i$. One option would be to immediately start decoding after $\tobs$ and discard the predictions in the offset between $\tobs$ and $\tfut$. However, auto-regressive decoding might lead to cascading errors over the offset. Instead, we address this one-to-many issue by injecting the offset information into $\bm{e}_\mathrm{ind}^i$. The decoder then receives a unique encoded representation for every $\tfut$ corresponding to the same $\tobs$. We do this by repurposing the idea of sinusoidal positional encodings \cite{vaswaniAttentionAllYou2017} to encode window offsets rather than relative token positions in sequences. For a given $\tobs$ and $\tfut$, and $d_e$-dim $\bm{e}_\mathrm{ind}^i$ we define the offset as $\Delta t = f1 - oT$, and the corresponding offset encoding $OE_{\Delta t}$ as
\begin{subequations}
\begin{gather}
    OE_{(\Delta t, 2m)} = \sin(\Delta t/10000^{2m/d_e}), \\
    OE_{(\Delta t, 2m+1)} = \cos(\Delta t/10000^{2m/d_e}).
\end{gather}
\end{subequations}
Here $m$ refers to the dimension index in the encoding. We finally compute the representation $\bm{e}^i$ for \autoref{eq:spdef} and \autoref{eq:aspdef} as
\begin{equation}
    \bm{e}^i = \bm{e}_\mathrm{ind}^i + OE_{\Delta t}.
\end{equation}

\textbf{Auxiliary Loss Functions}. We incorporate a geometric loss function for each of our sequence decoders to improve performance in pose regression tasks. For $\mathrm{p}_i$ at time $t$, given the ground truth $\bm{b}^i_t = [\bm{q};\bm{l};s]$, and
the predicted mean $\bm{\hat{b}}^i_t = [\bm{\hat{\bm{q}}};\bm{\hat{\bm{l}}};\hat{s}]$, we denote the tuple $(\bm{b}^i_t, \bm{\hat{b}}^i_t)$ as $B^i_t$. We then have the location loss in Euclidean space $\mathcal{L}_{\mathrm{l}}(B^i_t) = || \bm{l} - \bm{\hat{\bm{l}}} ||$, and we can regress the quaternion values using
\begin{equation}
    \mathcal{L}_{\mathrm{q}}(B^i_t) = \norm{\bm{q} - \frac{\hat{\bm{q}}}{\norm{\hat{\bm{q}}}}}.
\end{equation}
\citet{kendallGeometricLossFunctions2017} show how these losses can be combined using the homoscedastic uncertainties in position and orientation, $\hat{\sigma}^{2}_{\mathrm{l}}$ and $\hat{\sigma}^{2}_{\mathrm{q}}$:
\begin{equation}
    \mathcal{L}_{\sigma}(B^i_t) = \mathcal{L}_{\mathrm{l}}(B^i_t)\exp(-\hat{s}_{\mathrm{l}}) + \hat{s}_{\mathrm{l}}+\mathcal{L}_{\mathrm{q}}(B^i_t)\exp(-\hat{s}_{\mathrm{q}}) + \hat{s}_{\mathrm{q}},
\end{equation}
where $\hat{s} \coloneqq \log\hat{\sigma}^2$. Using the binary cross-entropy loss for speaking status $\mathcal{L}_{\mathrm{s}}(B^i_t)$, we have the overall auxiliary loss over $t \in \tfut$:
\begin{equation}
    \mathcal{L}_{\mathrm{aux}}(\bm{Y}, \bm{\hat{Y}}) = \sum_i \sum_t \mathcal{L}_{\sigma}(B^i_t) + \mathcal{L}_{\mathrm{s}}(B^i_t).
\end{equation}
The parameters of the SP and ASP are trained by maximizing the ELBO (\autoref{eq:elbo})
and minimizing this auxiliary loss.

\section{Experiments and Results}
\label{sec:experiments}
\subsection{Experimental Setup}

\textbf{Evaluation Metrics}. Prior forecasting formulations output a single future. However, since the future is not deterministic, we predict a future distribution. Consequently, needing a metric that accounts for probabilistic predictions, we report the log-likelihood (LL) $\log p(\bm{Y}|\bm{X},C)$, commonly used by all variants within the NP family \cite{garneloNeuralProcesses2018, kimAttentiveNeuralProcesses2019, singhSequentialNeuralProcesses2019}. The metric is equal to the log of the predicted density evaluated at the ground-truth value. (Note: the fact that the vast majority of forecasting works even in pedestrian settings omit a probabilistic metric, using only geometric metrics, is a limitation also observed by \citet[Sec.~8.3]{rudenko2020human}.) Nevertheless, for additional insight beyond the LL, we also report the errors in the predicted means\textemdash geometric errors for pose and accuracy for speaking status\textemdash and provide qualitative visualizations of forecasts.

\textbf{Models and Baselines}. In keeping with the task requirements and for fair evaluation, we require that all models we compare against forecast a distribution over future cues. 
\begin{itemize}[itemsep=0em]
\item To evaluate our core idea of viewing conversing groups as meta-learning tasks, we compare against non-meta-learning methods: we adapt variational encoder-decoder (VED) architectures \cite{haNeuralRepresentationSketch2017, bowmanGeneratingSentencesContinuous2016} to output a distribution. 
\item To evaluate our specific modeling choices within the meta-learning family, we compare against the NP and ANP models (see Section~\ref{sec:methodology}). The original methods were not proposed for sequences, so we adapt them by collapsing the timestep and feature dimensions in the data.
\end{itemize}
Note that in contrast to the SP models, these baselines have direct access to the future sequences in the context, and therefore constitute a strong baseline.
We consider two variants for both NP and SP models: \textit{-latent} denoting only the stochastic path; and \textit{-uniform} containing both the deterministic and stochastic paths with uniform attention over context sequences. We further consider two attention mechanisms for the cross-attention module: \textit{-dot} with dot attention, and \textit{-mh} with wide multi-head attention \cite{kimAttentiveNeuralProcesses2019}.
Finally, we experiment with two choices of backbone architectures: multi-layer perceptrons (MLP), and Gated Recurrent Units (GRU). Code, processed data, trained models, and test batches for reproduction are available at {\small \url{https://github.com/chiragraman/social-processes}}.

\subsection{Evaluation on Synthetic Data}

\textbf{Synthesising Two Single-Modality Datasets.} To validate our method on simplified social settings, we synthesize two datasets, simulating separate modalities -- head movements and speaking turn changes. These two social modalities are extensively studied in psychology research \cite{argyle1994gaze, sacksSimplestSystematicsOrganization1974}, as they provide insights into the dynamics of human interactions. In particular, head movement has been shown to influence conversational flow and engagement, as well  as convey emotional states and reinforce verbal communication \cite{mcclave2000linguistic, kendon1972some}. 
Similarly, speaking turn-taking dynamics play a big role in structuring social interactions, supporting smooth exchanges and effective communication \cite{duncanSignalsRulesTaking1972}. 
Turn-taking has also been extensively analyzed in studies of conversational agents and collaborative systems \cite{skantze2021turn, jokinen2013gaze, Raux2012-fj}.

Therefore, our synthetic datasets model fundamental social behaviors, making them valuable for testing the applicability and adaptability of our method. Also, their simplicity allows us to analyze important properties of the proposed models, such as generalization and latent space dynamics without losing connection to realistic scenarios.




\textbf{First Dataset -- Glancing Behaviour.} We synthesize a dataset simulating two glancing behaviors in social settings \cite{mooreNonverbalCourtshipPatterns1985}, approximated by horizontal head rotation. The sweeping \textit{Type I} glance is represented by a 1D sinusoid over $20$ timesteps. The gaze-fixating \textit{Type III} glance is denoted by clipping the amplitude for the last six timesteps. The task is to forecast the signal over the last $10$ timesteps ($\tfut$) by observing the first $10$ ($\tobs$). Consequently, the first half of $\tfut$ is certain, while the last half is uncertain: every observed sinusoid has two ground truth futures in the data (clipped and unclipped). It is impossible to infer from an observed sequence alone if the head rotation will stop partway through the future. Based on this raw dataset, we design two meta-datasets\footnote{from now on, we refer to \textit{meta-datasets} simply as \textit{datasets}} which differ only in how sequences are assigned to meta-samples. We call the datasets \textit{mixed context} and \textit{separated context}. In \textit{mixed context}, every meta-sample contains sequences of both types of glances, while in \textit{separated context}, every meta-sample contains only one type of sequence -- some meta-samples are \textit{Type I} and some are \textit{Type III}. As a result, in the meta-samples from \textit{mixed context} dataset it is impossible to infer if the future is clipped or not, while in the \textit{separated context} dataset, the future can be inferred exactly, based on the provided context. When training models on these two datasets, we use a $1$-dimensional latent variable, as the necessary latent information is inherently binary -- either all underlying sequences are Type I or Type III.


\textbf{Second Dataset -- Speaking Turns.}
We further synthesize interactions comprised only of speaking-turn information. In particular, we introduce four new datasets each simulating $5$ people having a conversation while standing in a circle. We consider two types of group dynamics:
\begin{itemize}
    \item \textit{Free-for-all interactions}, where the group conversation is balanced -- all participants talk approximately the same amount of time. For this category we create 3 separate datasets with increasing levels of freedom in a conversation:
    \begin{itemize}
        \item \textit{Dual interactions}, simulating a restricted social setting where after a person finishes speaking, the next speaker is always a person to their left or is always a person to their right, depending on the group. As a result, the groups can be characterized as "clockwise", or "anticlockwise".
        \item \textit{Dual-random interactions}, simulating a social setting where after a person finishes speaking, the next speaker can arbitrarily be either the person to their left or a person to their right.
        \item \textit{Full-random interactions}, simulating a free social setting where after a person finishes speaking, the next speaker can be anybody else in the group.

    \end{itemize}
    
    \item \textit {Dominated interactions}, where a single individual in a group is taking control of the conversation by interjecting between every other speaker.
    For this setting we create a single dataset where conversationalists take turns speaking either clockwise or anti-clockwise, depending on the group. However, instead of the conversation being a round-the-circle interaction, the dominating person always talks every two turns. As a result, every group can be identified by two latent features: the (index of) dominating person and the direction of conversation (clockwise or anti-clockwise).
\end{itemize}

In all four datasets we construct the meta-samples in the same way. Each meta-sample contains interactions from a single group from which we sample 8 sequences for the context and 11 sequences for the target. Each sequence in a context contains 6 observed time steps, and each sequence in a target contains 2 observed time steps and 4 future time steps that need to be predicted. In all datasets, each speaking turn lasts exactly two timesteps. Furthermore, an updated version of the SP model produces a categorical distribution (of size $5$) for each timestep, corresponding to the probability of each person speaking at that time. The loss function does not change as we still try to minimize the NLL, just the evaluated distribution is categorical instead of Gaussian. We use RNN encoders and decoders and have a $64$-dimensional latent space.

\begin{figure}[!t]
\centering
\begin{minipage}[t]{\columnwidth}
  \centering
  \captionsetup{type=figure}
  \includegraphics[width=0.9\textwidth]{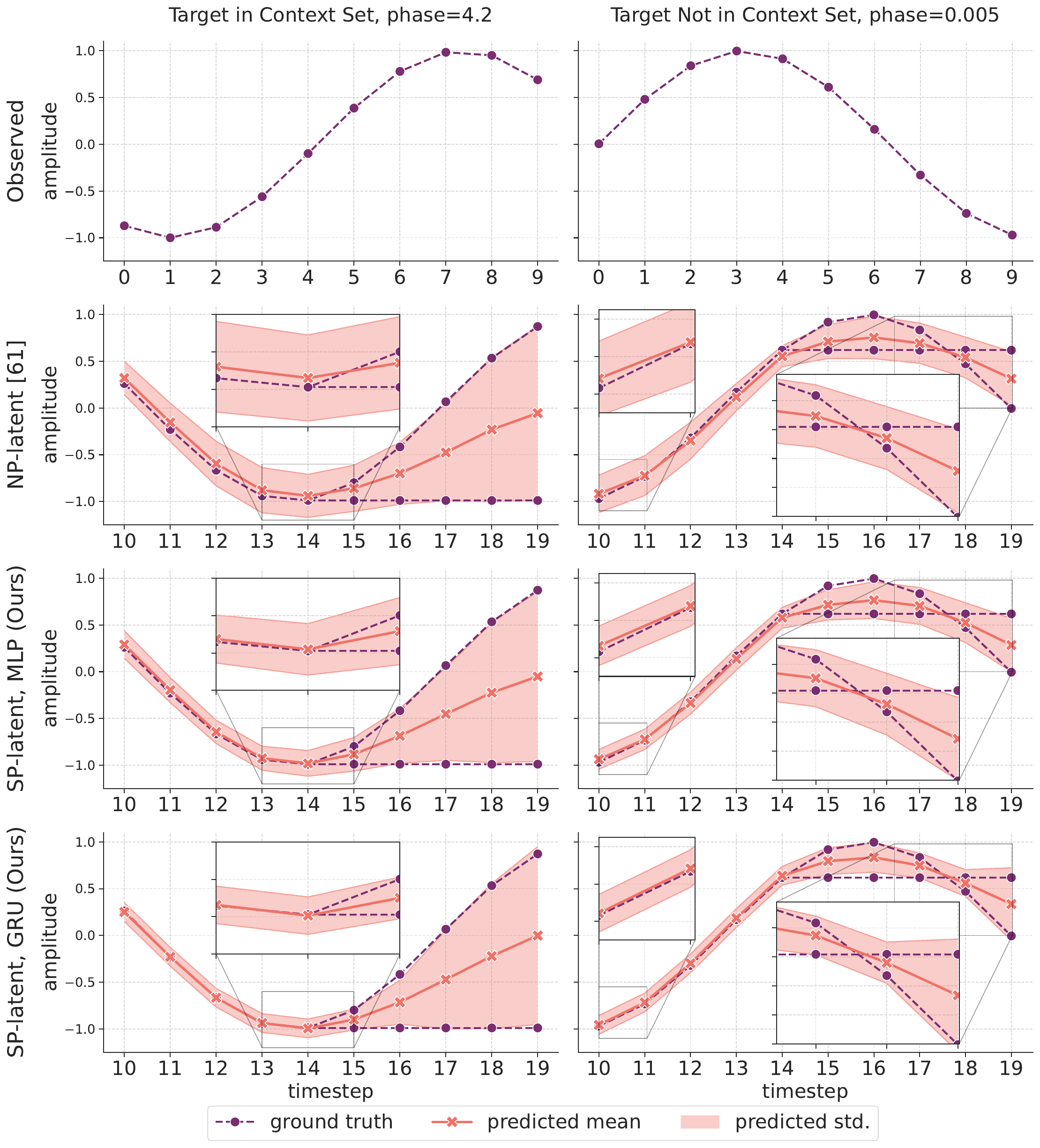}
  \captionof{figure}{Ground truths and predictions for the \textit{mixed context} glancing behavior task. All models learn to average over the possible futures. Our SP models learn a better fit than the NP model, SP-GRU being the best (see zoomed insets).}
  \label{fig:toy}
\end{minipage}
\end{figure}

\begin{figure}[!t]
    \centering
      \includegraphics[width=0.75\columnwidth]{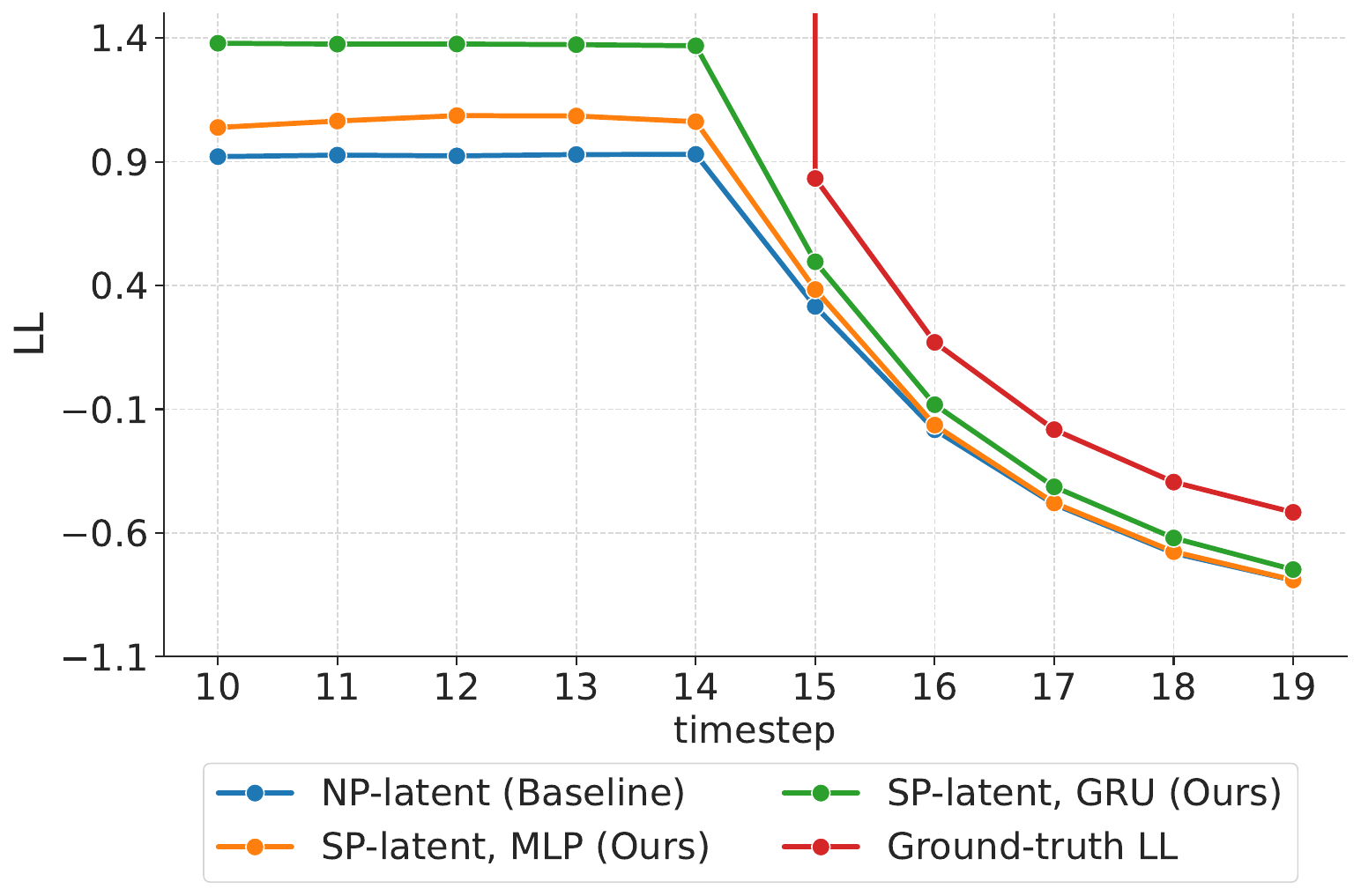}
      \caption{Mean per timestep LL over the sequences in the synthetic glancing \textit{mixed context} dataset. Higher is better.}
    \label{fig:ts-synthetic-ll}
\end{figure}

\textbf{Accuracy: SP outperforms the baseline}. We first compare the performances of different models on the synthetic glancing behaviour dataset. \autoref{fig:toy} illustrates the predictions for two sample sequences, trained and tested on \textit{mixed context} dataset. Table~\ref{tab:glancing-metrics} provides quantitative metrics and \autoref{fig:ts-synthetic-ll} plots the LL per timestep. The LL is expected to decrease over timesteps where ground-truth futures diverge, being $\infty$ when the future is certain. We observe that all models estimate the mean reasonably well, although our proposed SP models perform best. More crucially, the SP models, especially the SP-GRU, learn much better uncertainty estimates compared to the NP baseline (see zoomed regions in \autoref{fig:toy}). We provide additional analysis, and data synthesis details in \Cref{app:results,app:data} respectively.

\begin{figure}[!t]
    \centering
    \captionsetup{type=table}

\input{synthetic_metrics}

\end{figure}


\begin{figure*}[!t]
    \centering
    \begin{subfigure}{0.45\textwidth}
        \centering
        \includegraphics[width=0.97\textwidth]{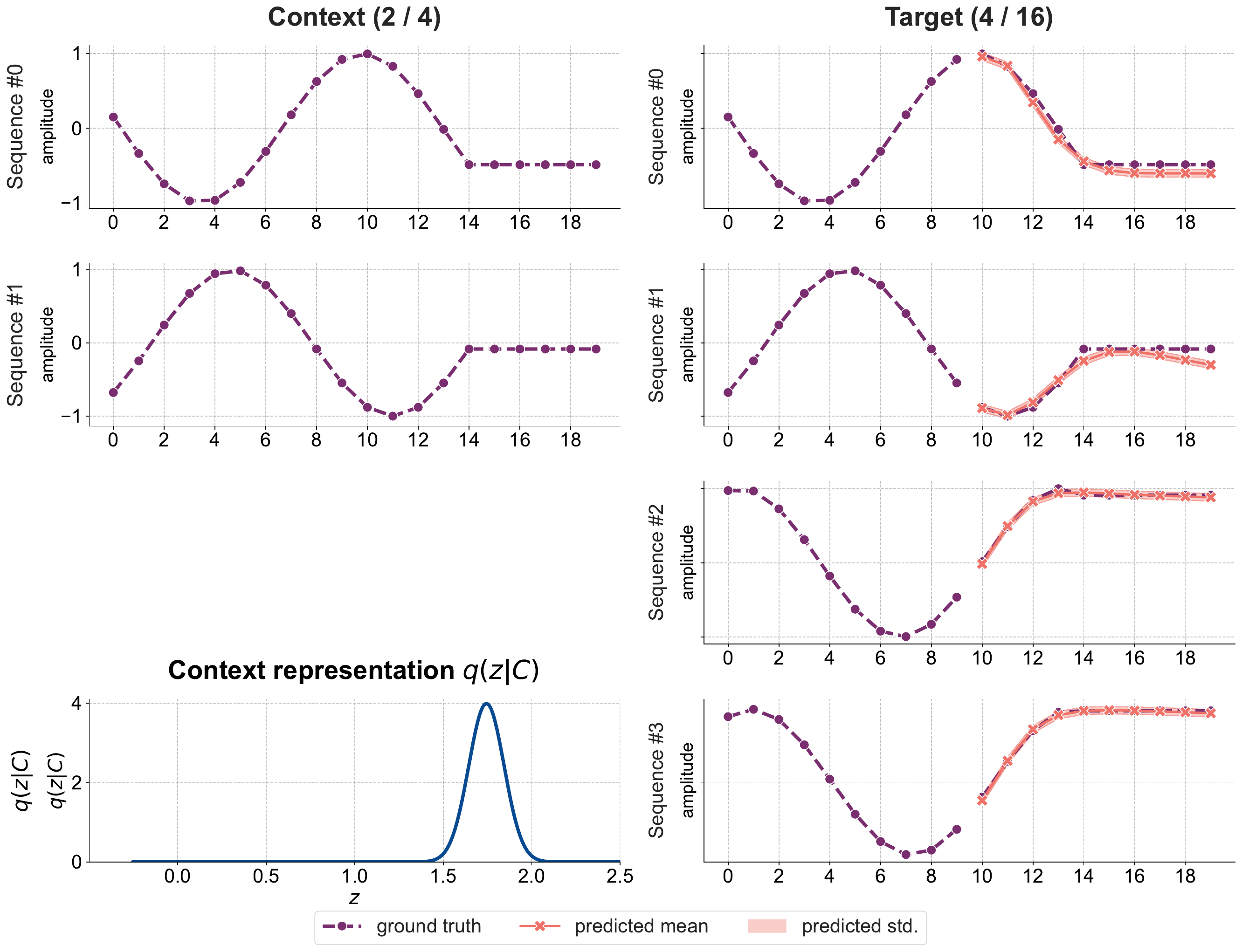}
        \caption{An instance of a \textit{Type I} meta sample}
    \end{subfigure} \hspace{0.01\textwidth} 
    \begin{subfigure}{0.45\textwidth}
        \centering
        \includegraphics[width=0.97\textwidth]{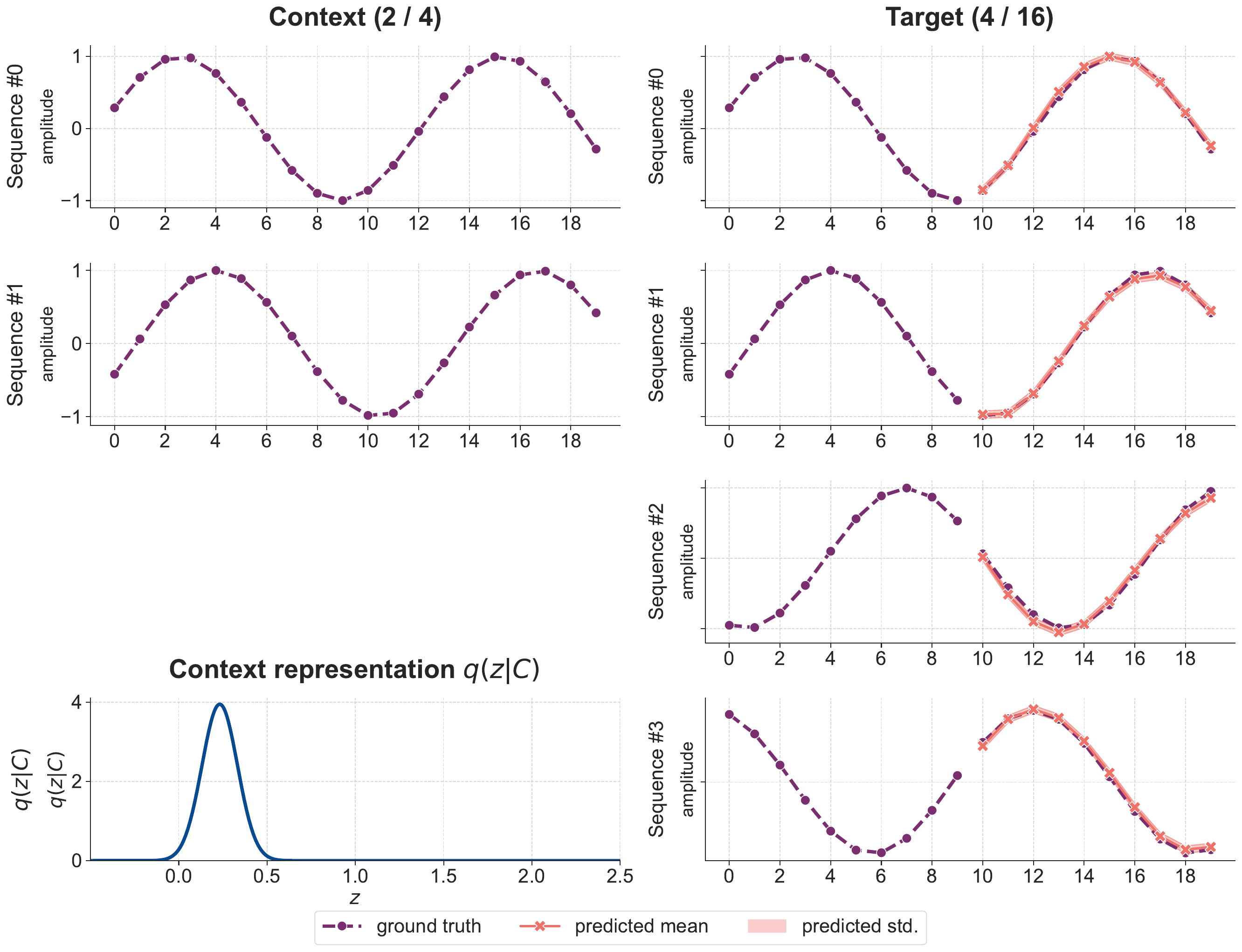}
        \caption{An instance of a \textit{Type III} meta sample}
    \end{subfigure}
  \captionof{figure}{Two different meta samples, representing synthetic \textit{Type I} and \textit{Type III} glances. Both subfigures contain two columns: one for the context and one for the target. The context column depicts 2 sequences (out of 4) and the context representation  $q(z|C)$. We observe that the contexts of \textit{Type I} and \textit{Type III} meta samples get mapped to distant distributions: \textit{Type I} contexts gets mapped to a Normal distribution with mean of approx. $0.25$ and \textit{Type III} contexts get mapped to Normal distributions of mean $1.75$, which suggests that the model has learned to differentiate between different types of glancing behaviours. Finally, in the second columns of both subfigures, 4 (out of 16) target sequences are depicted, together with the model predictions. Here we observe that the model is able to predict the glancing behaviour almost perfectly.}
  
  \label{fig:toy-simple-predictions}

\end{figure*}

\textbf{Latent Space: Posterior Collapse Can Be Avoided with Informative Contexts.} As seen in \autoref{fig:toy}, when trained on the \textit{mixed context} dataset, our model learns to average between the two possible futures, as it is not provided enough information to infer the sequence type. As a result, we find that posterior collapse \cite{lucas2019understanding, kingma_2013_autoencoding} is occurring -- the sequence encoder maps all contexts from the \textit{mixed context setting} to the same distribution $q(z | C)$. 

\begin{figure*}[!t]
    \centering
    \begin{subfigure}[t]{0.234\textwidth}
        \centering
        \includegraphics[width=\textwidth]{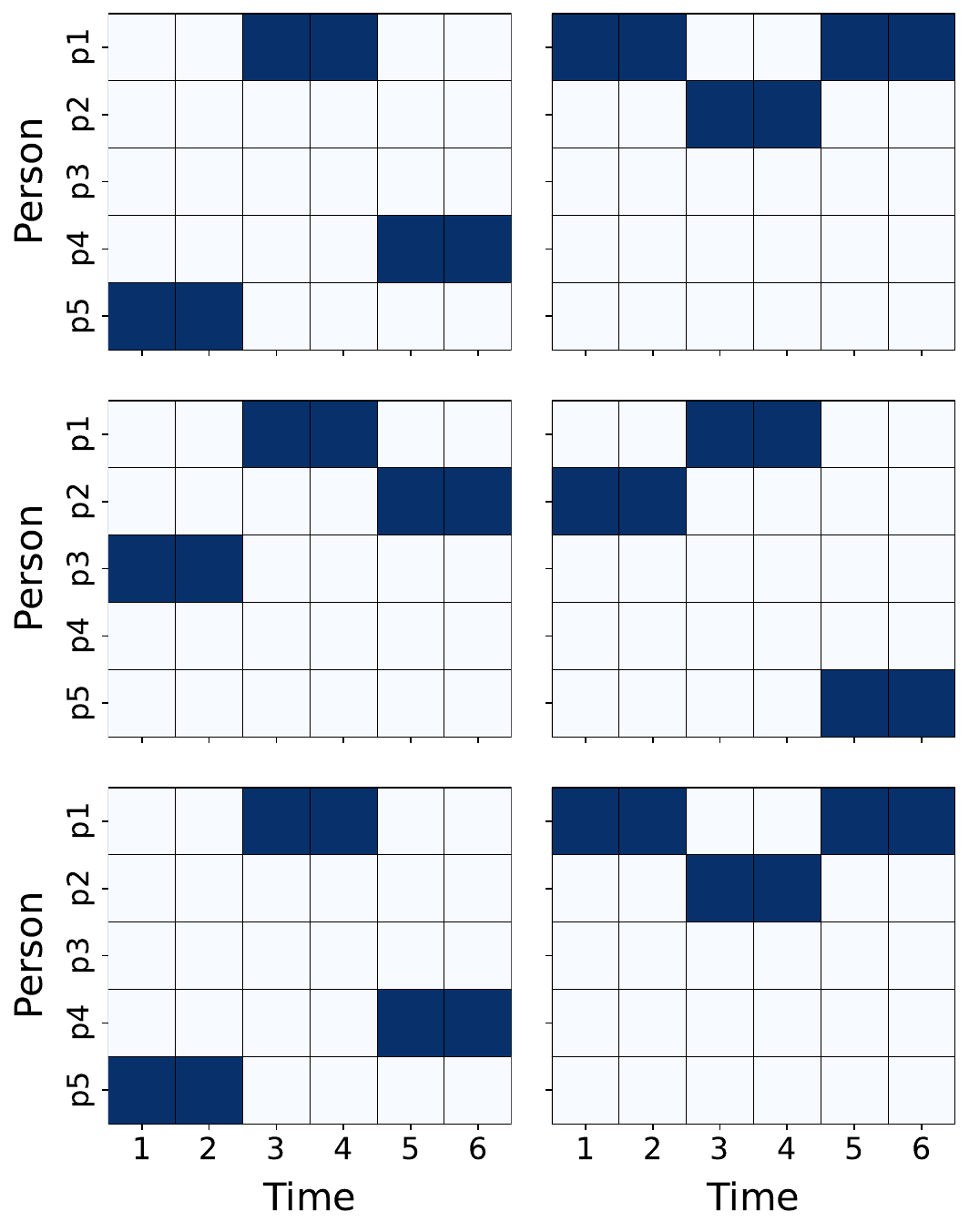}
        \caption{\textbf{Context}. An example of context that was provided to the models (we show 6 out of 8 sequences here). This same context was provided to all 3 models in this figure (b, c and d). Each grid corresponds to one sequence -- the dark-blue squares denote the speaker at the current moment. It can be deduced from the context figure that the dominating person is the first person, corresponding to row 1, as the first speaker talks in between all others.}
    \end{subfigure}
    \hspace{0.15cm}
    \begin{subfigure}[t]{.234\textwidth}
        \centering
        \includegraphics[width=\textwidth]{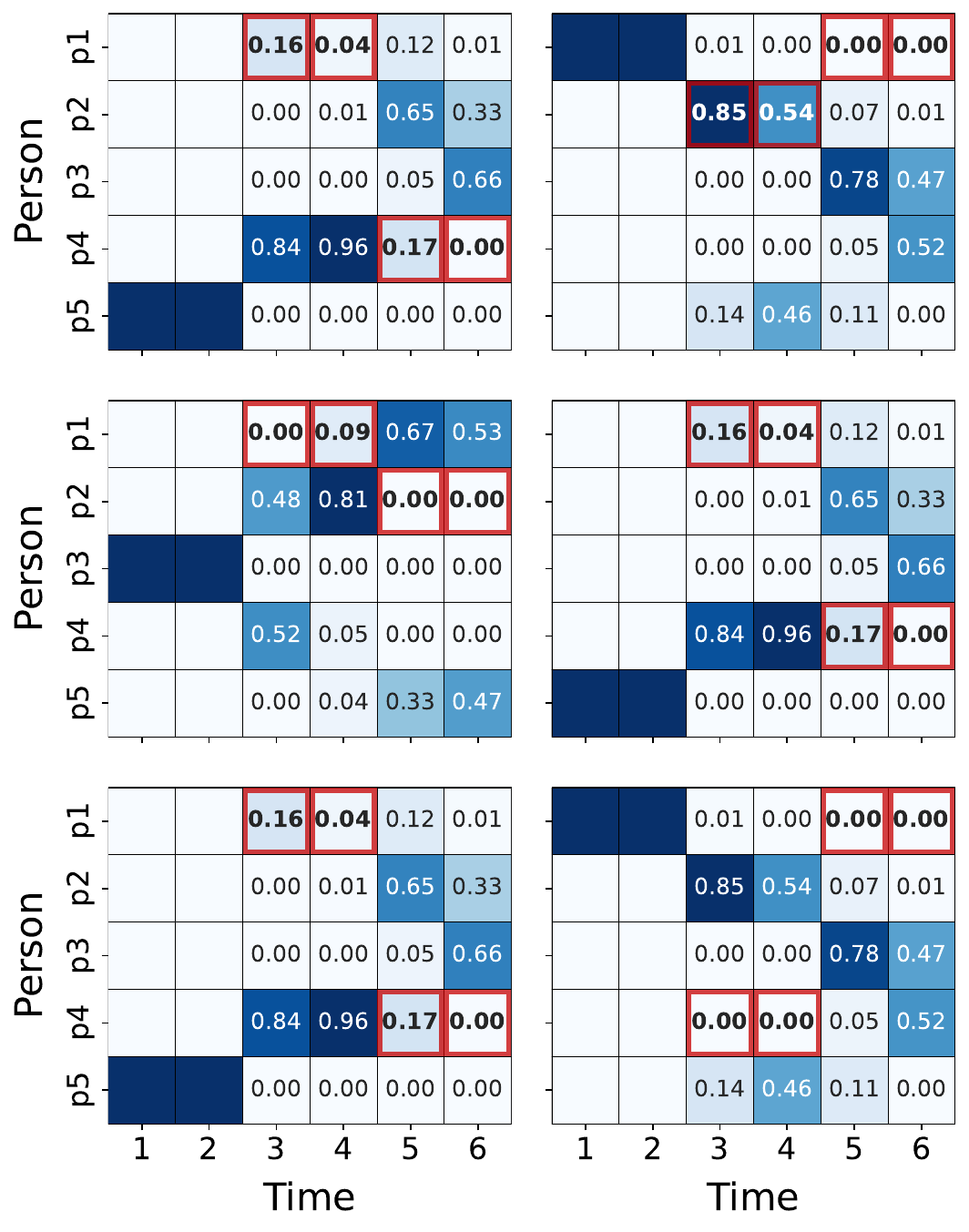}
        \caption{\textbf{Dual model.} An example of predicted futures with a model trained on the Dual dataset. In particular, we observe that the model outputs some averaged version of the 2 types of futures it was trained on, which does not fit in with the context at all. Note that this is only a single example. We find that in different meta samples, the model sometimes predicts a single future with high certainty, instead of having the two branches as visualized here.}
    \end{subfigure}
    \hspace{0.15cm}
    \begin{subfigure}[t]{.234\textwidth}
        \centering
        \includegraphics[width=\textwidth]{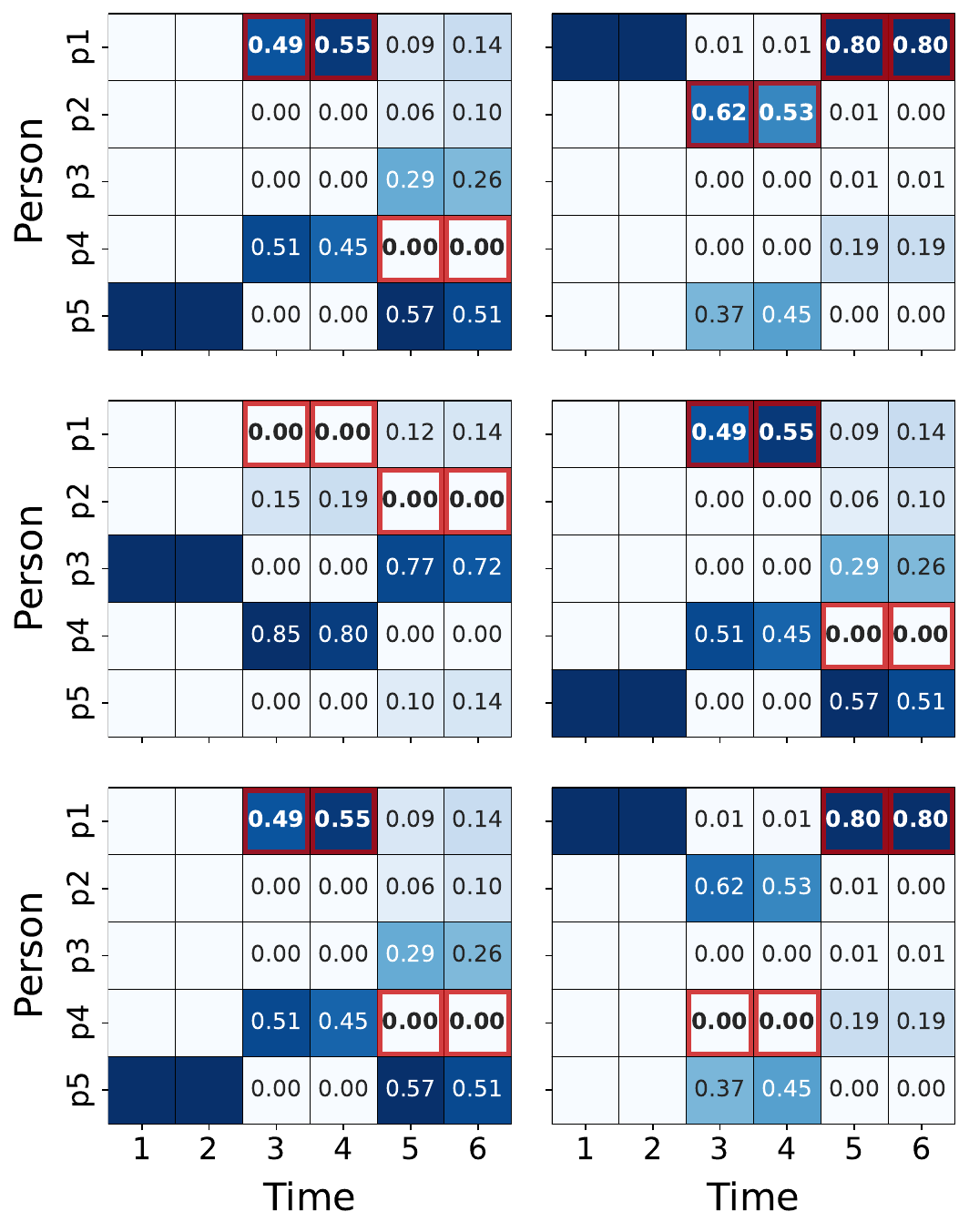}
        \caption{\textbf{Dual-random model}. An example of predicted futures with a model trained on the Dual-random dataset. We can see that the model does not adapt to the context, and instead outputs a checkers-like pattern, as it is the exact pattern that would minimize the loss for the \textit{Dual-random} dataset.}
        \label{fig:speaking_predictions_c}

    \end{subfigure} 
    \hspace{0.15cm}
    \begin{subfigure}[t]{.234\textwidth}
        \centering
        \includegraphics[width=\textwidth]{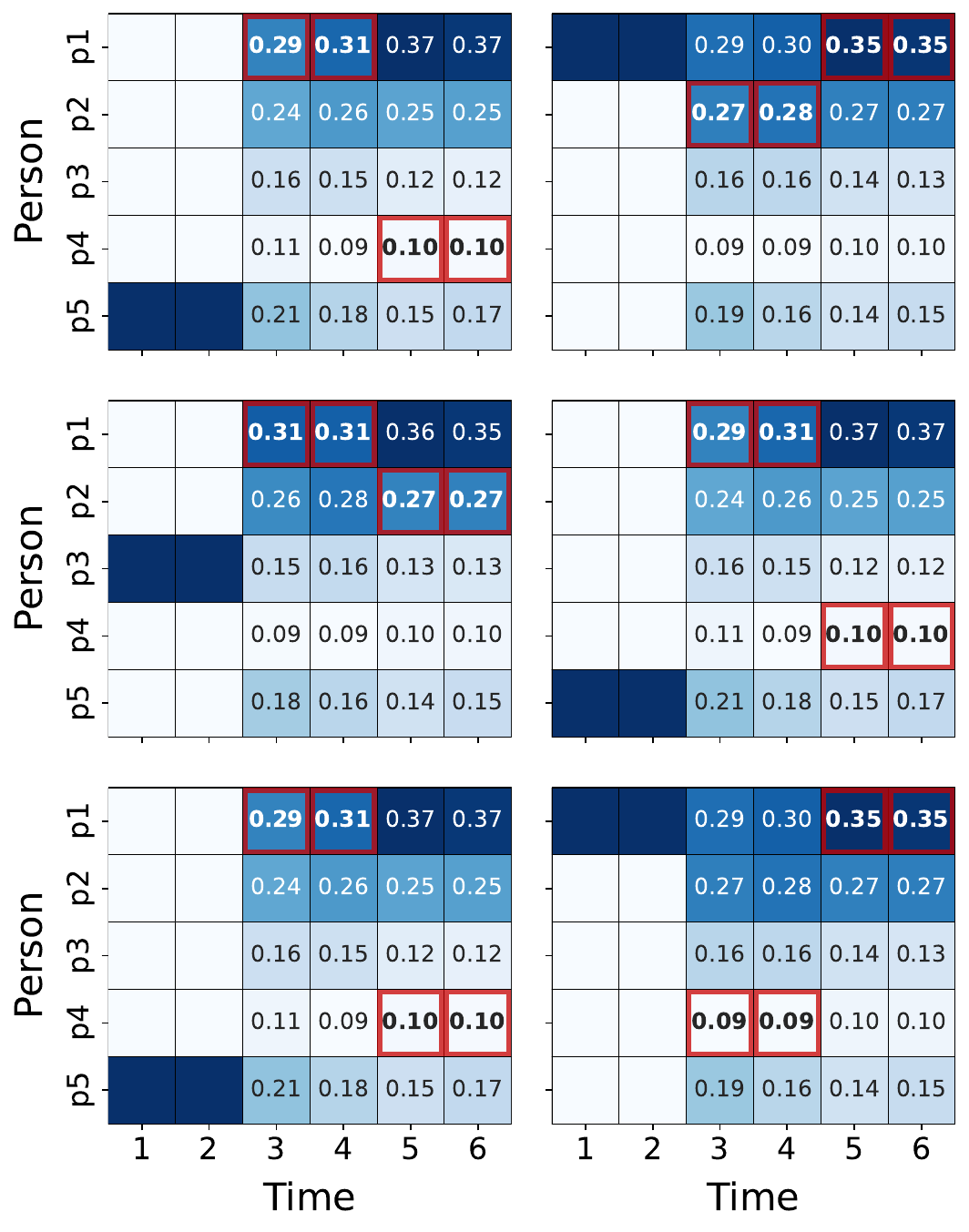}
        \caption{\textbf{Full-random model.} An example of predicted futures with a model trained on the Full-random dataset. We observe that the model takes the context into account, as the highest speaking probability at every timestep is assigned to person 1, who dominates the conversation. However, the model does not differentiate probabilities for different timesteps --- roughly the same probability is assigned to each person for all timesteps. However, this shows that model is extracting information from the context.}
    \end{subfigure}

  \captionof{figure}{Comparison of three models trained on the Dual, Dual-random and Full-random datasets. Dark blue cells denote the observed sequences provided to the models, while the brighter blue cells denote the model's predicted sequences and the number inside of each cell represents the model's output \textemdash predicted probability of a person being a speaker at a given time. The darker the cell, the more probability mass the model assigned to a person at a given timestep. The cells with a red s denote the ground truth future. }
    \label{fig:speaking_predictions}

\end{figure*}

To investigate the model's proclivity to collapse the posterior, we further test it on datasets with more useful information in the contexts. First, we train the model on synthetic glancing \textit{separated context} dataset, and find that it learns to differentiate between the two possible futures based on the context. The model achieves that by mapping the two different context types to very distant latent distributions (see \autoref{fig:toy-simple-predictions}), so the posterior collapse does not occur. We also train the model on the \textit{Dual}, \textit{Dual-random} and \textit{Full-random} datasets and after hyperparameter tuning we find that full posterior collapse is not occurring with any of the datasets. In particular, the models map different contexts to different latent points, and different latent points result in generally different outputs. It must also be noted that the absence of posterior collapse does not necessarily result in a large variance in the output space. For instance, for any provided context, the model trained on the \textit{Dual-random} dataset always predicts a very similar pattern as its output -- a checkers-like pattern (see \autoref{fig:speaking_predictions_c}) that maximizes LL under the uncertainty present in the \textit{Dual-random} dataset.

\begin{figure}[!t]
    \centering
    \includegraphics[width=1\columnwidth]{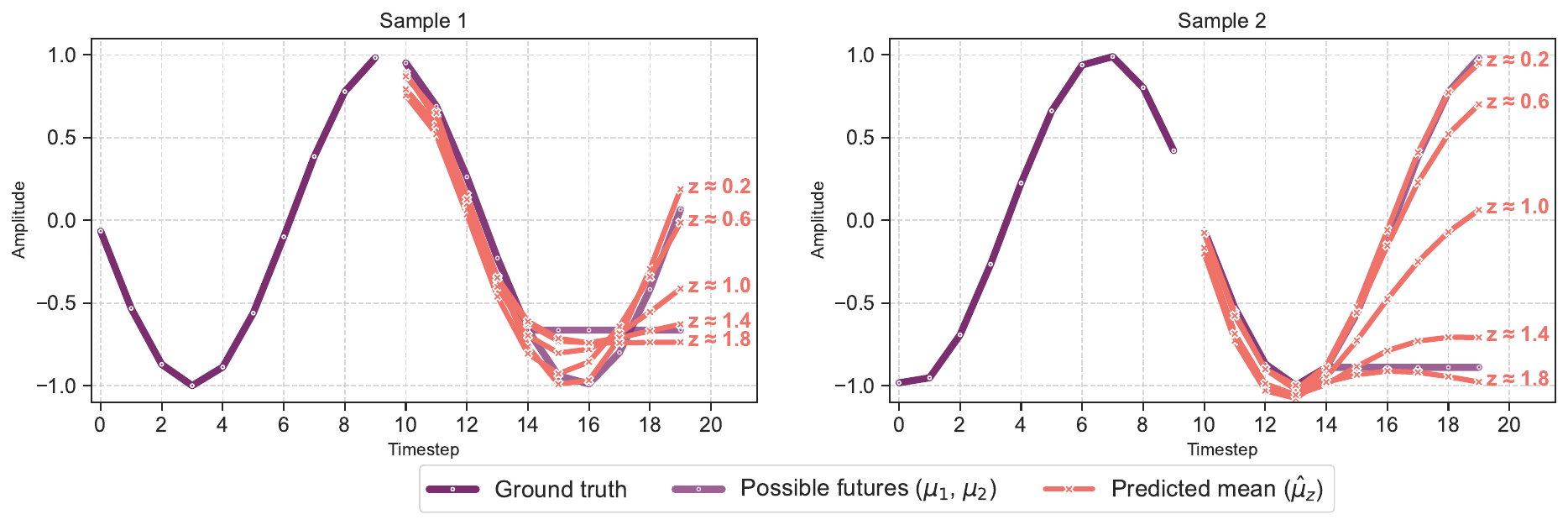}
    \captionof{figure} {A visual representation of latent space in the synthetic glancing experiment. Instead of encoding some context to $z$, we sample $z$ uniformly from the interval $[0.25; 1.75]$ (the ends of the interval represent \textit{Type I} and \textit{Type III} contexts respectively). We inject these values into 2 different observed sequences (Samples 1 and 2) and observe that as the $z$ value increases from $0.25$ to $1.75$, the outputs smoothly transition from \textit{Type I} to \textit{Type III}. For instance, the latent mid-point $z=1$ corresponds to the prediction, which is also in the middle between the two possible futures. This further suggests that the SP model has learned a useful latent representation of the data. }
    \label{fig:toy-z-analysis}
\end{figure}

\begin{figure}[!t]
    \centering
        \centering
        \includegraphics[width=\columnwidth]{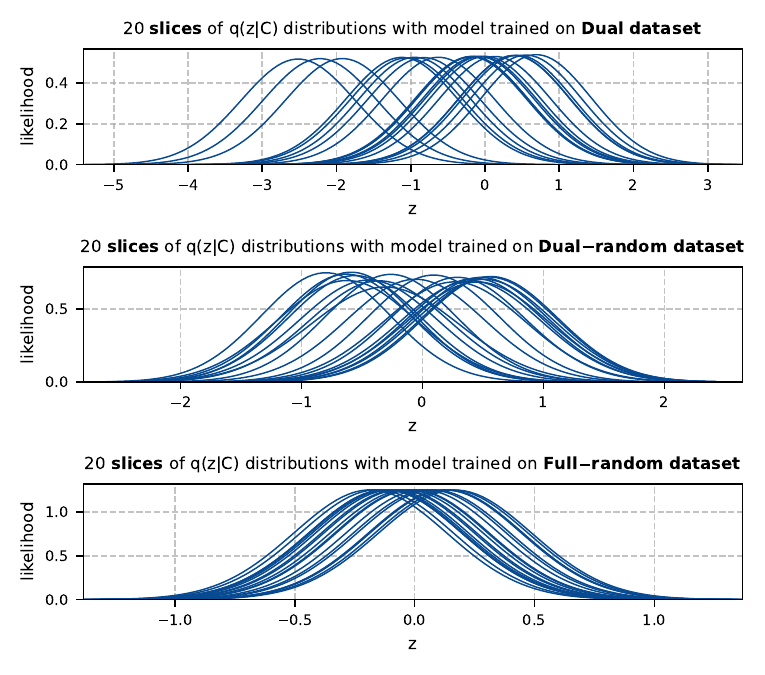}
        \caption{Encodings of different contexts with 3 different models. Each plot shows 20 different curves -- each of them represent an encoding of some meta sample's context. The meta samples are taken from the Dominating dataset. This allows looking into the latent space in order to inspect whether posterior collapse is occurring. Note that $q(z|C)$ is a $64$-dimensional Gaussian, but we are plotting only a single-dimensional slice.}
  
  \label{fig:speaking_posteriors}

\end{figure}

\textbf{Latent Space: SP Models Can Learn a Semantic Latent Mapping.} We find that the model learns a semantic representation of the latent space when trained on the \textit{separated context} glancing behaviour dataset. As mentioned before, it learns to map the two different context types to two distant distributions $q(z|C)$. In particular, the contexts containing \textit{Type I} glancing sequences get mapped to a Normal distribution with mean approximately $0.25$ and the contexts with \textit{Type III} glancing sequences get mapped to mean $1.75$ (see \autoref{fig:toy-simple-predictions}). Furthermore, \autoref{fig:toy-z-analysis} shows how other $z$ values in the interval $[0.25, 1.75]$ surprisingly represent futures "in-between" the \textit{Type I} and \textit{Type III}: they represent futures where the head rotation neither fully stops, neither fully continues. The closer $z$ is to $1.75$, the slower head rotation continues. Therefore, even when the model was trained on two extreme cases (head movement fully stopping, or continuing), it has learned to interpolate between them in the latent space. That suggests that the SP models in general could be capable of learning useful latent representations of contextual data.
We also find indications of the model's ability to map to latent space by inspecting how the encoder maps out of distribution samples to the latent space. In particular, we take the 3 models, trained on each of the \textit{Free-for-all} datasets, and for each of them, map contexts from the \textit{Dominating} dataset, and observe where these contexts end up (see \autoref{fig:speaking_posteriors}). We find that different contexts get mapped to different positions in the latent space even for these unseen contexts. This does not imply that the mappings are meaningful, however, it allows again concluding that the posterior does not collapse and that differentiation is happening in the latent space.

\textbf{Generalization: Models Can Interpolate But Not Extrapolate.} In the \textit{separated context} glancing experiment we saw a good indication that SP models can generalize to unseen data, by interpolating between the two extreme values the model was trained on. We further investigate capabilities for stronger generalization, by taking models, trained on \textit{Free-for-all} datasets and seeing if they can generalize for the \textit{Dominating} dataset. In particular, we aim to figure out how different training settings impact generalization performance, as the \textit{Dual}, \textit{Dual-random} and \textit{Full-random} datasets differ regarding how much varied the social behaviour presented to the model is. As can be observed in \autoref{tab:speaking_losses}, better testing loss values were achieved with more general datasets: the dataset that contained the most varied behaviour (i.e., the \textit{Full-random} dataset) resulted in the lowest loss, while the dataset with the least varied behaviour (i.e., the \textit{Dual} dataset) produced a model which resulted in the highest loss. 

However, loss alone does not paint the full picture. By examining the outputs produced by the models when tested on the \textit{Dominating} dataset (\autoref{fig:speaking_predictions}), we find that the \textit{Dual} and \textit{Dual-random} models do not generalize to the new dataset, as both of them predict the same type of behaviours they were trained on. However, the model trained on \textit{Full-random} dataset performs significantly better -- it uses the information from the context to identify the dominating person and then assigns the most probability mass to this person across all future time steps. Such strategy is suboptimal, nevertheless, it shows generalization, since the model takes unseen dynamics into account.

The \textit{Full-random} model's success with the \textit{Dominating} dataset can partly be explained by the fact that the social dynamics in the \textit{Dominating} dataset are effectively a subset of those in the \textit{Full-random} dataset. This suggests that the model only had to interpolate in order to generalize for the unseen \textit{Dominating} dataset. On the other hand, the \textit{Dual} and \textit{Dual-random} models were unable to generalize to the \textit{Dominating} dataset, likely because their training data included only one type of behaviour, which was not similar to those in the \textit{Dominating} dataset.

We conclude that the SP model's ability to generalize is limited by the variety of social behaviour types in the training set. The model is able to interpolate between known social behaviors, as demonstrated by the \textit{Dual} model producing averaged behaviors, the \textit{Full-random} model performing well on the \textit{Dominating} dataset and the \textit{separated context} experiment. However, the SP model has difficulty extrapolating to out-of-distribution data. Therefore, we could expect the model to generalize to unseen behaviours provided that the training dataset includes social interactions taken from a similar distribution.

\begin{table}[t]
    \centering
    \ra{1.1}
    \setlength{\tabcolsep}{5pt}
    \centering
    \scriptsize
    \caption{Loss values evaluated on the \textit{Dominating} dataset for 3 models which were obtained by training on the \textit{Dual}, \textit{Dual-random} and \textit{Full-random} datasets. 
    \\ }    
    \begin{tabular}[b]{@{}lc@{}}
    \toprule
        \multirow{1}{*}{\textbf{Train dataset}} & \textbf{Loss} \\

    \midrule
      Dual  & 357.06 \\
      
      Dual-random  & 154.95  \\
    
     Full-random  & \textbf{65.15} \\
    \bottomrule
    \end{tabular}

    \label{tab:speaking_losses}

\end{table}

\newcommand{\rulesep}{\unskip\ \vrule\ }

\section{Discussion}
\label{sec:discussion}
Despite recent advances in low-level behaviour forecasting of social conversations, the field still remains largely underexplored.  
The setting of social conversations remains a uniquely challenging frontier for state-of-the-art low-level behavior forecasting. 
The predominant focus of researchers working on social human-motion prediction has been pedestrian trajectories \cite{rudenko2020human} or actions such as \textit{punching, kicking, gathering, chasing, etc.} \cite{yaoMultipleGranularityGroup2018, adeli2020socially}. In contrast to such activities which involve pronounced movements, the postural adaptation for regulating conversations is far more subtle. At the same time, the social intelligence required to understand the underlying dynamics that drive a conversation is comparatively more sophisticated than for an action such as a kick. We hope that the social-science considerations informing the design of SCF (joint probabilistic forecasting for all members) and the SP models (groups as meta-learning tasks) constitute a meaningful foundation for future research in this space to build upon. Note that for our task formulation, even the performance of our baseline models constitutes new results.

\textbf{Cross-Discipline Impact and Ethical Considerations}. While our work here is an \textit{upstream} methodological contribution, the focus on human behavior entails ethical considerations for downstream applications. One such application involves assisting social scientists in developing predictive hypotheses for specific behaviors by examining model predictions.
In these cases, such hypotheses must be verified in subsequent controlled experiments. With the continued targeted development of techniques for recording social behavior in the wild \cite{raman2020modular}, evaluating forecasting models in varied interaction settings would also provide further insight. 
Another application involves helping conversational agents achieve smoother interactions.
Here researchers should be careful that the ability to forecast does not result in nefarious manipulation of user behavior.


{\small
\bibliographystyle{unsrtnat}
\bibliography{references}
}

\appendices
\input{appendix}

\end{document}

%% file: figures/concept.tex
\begin{figure}[t!]
\makebox[\columnwidth][c]{\includegraphics[width=\columnwidth]{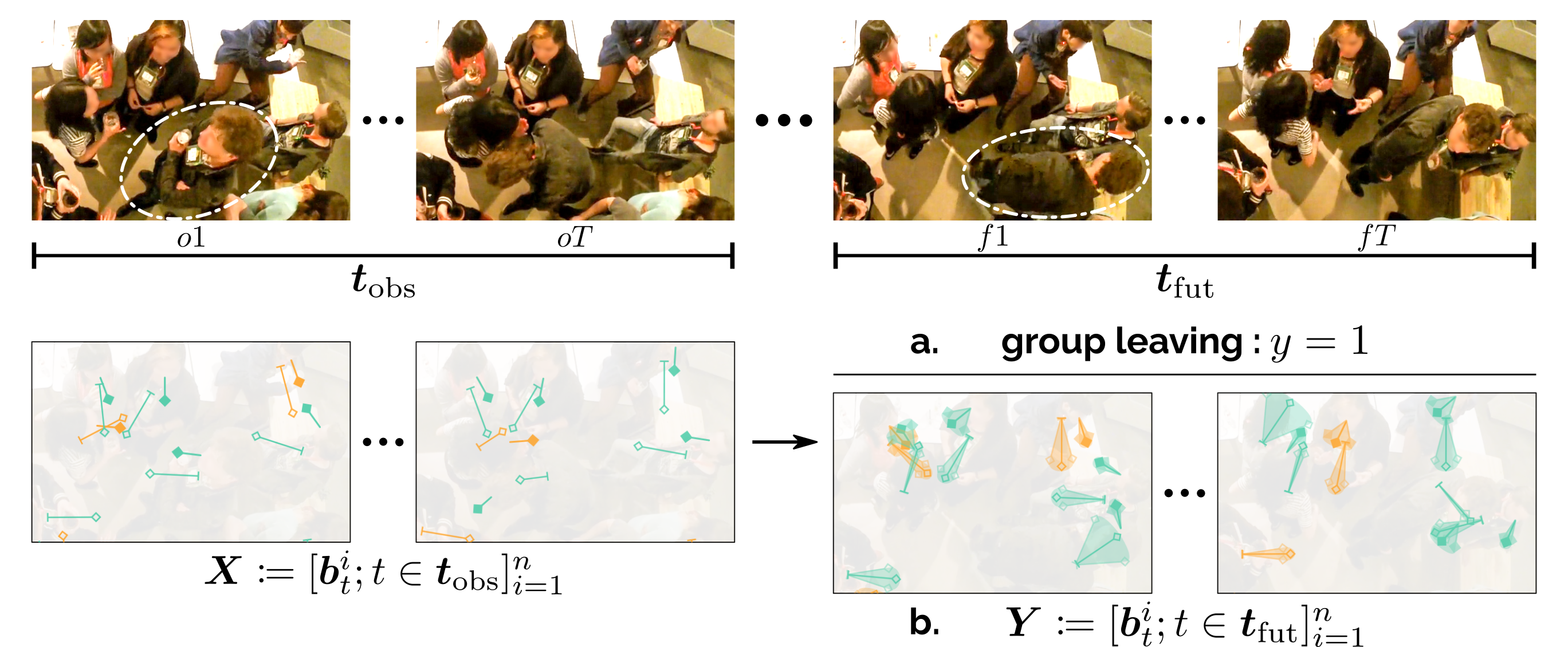}}
\caption{Illustration of the two forecasting approaches on a real-world situation from the MatchNMingle dataset \cite{cabrera2018matchnmingle}. The top part of the figure illustrates a high-order \textit{group leaving} event \cite{vandoornRitualsLeavingPredictive2018}, where the individual leaves from one group (in $\tobs$) to another ($\tfut$). The bottom part depicts the low-level social cues $\bm{b}^i_t$: head pose (solid normal), body pose (hollow normal), and speaking status (speaker in orange), which are used as features for predictions. In the case of \textbf{top-down approach (a)}, the goal is to predict the group leaving label, therefore in this case, from 90 minutes of interaction, only 200 samples can be generated \cite{vandoornRitualsLeavingPredictive2018}. However, in the case of our proposed \textbf{bottom-up}, self-supervised formulation of \textit{Social Cue Forecasting} \textbf{(b)}, the task is to predict the future low-level cues. This allows to make use of all 90 minutes of data. 
}
\label{fig:concept}
\end{figure}

%% file: synthetic_metrics.tex
\ra{1.1}
\setlength{\tabcolsep}{2.5pt}
\centering
\scriptsize
\caption{\textbf{Mean (Std.) Metrics on the Synthetic Glancing Behavior Dataset.}. All models are \textit{latent} variants. The metrics are averaged over timesteps; mean and std. are then computed over sequences. Higher is better for LL, lower for MAE.}
\label{tab:glancing-metrics}
\resizebox{\columnwidth}{!}{%
\begin{tabular}[b]{@{}lcccccc@{}}
\toprule
    &   \multicolumn{2}{c}{Mixed context} & \multicolumn{2}{c}{Type I context} & \multicolumn{2}{c}{Type III Context} \\

    \cmidrule(lr){2-3} \cmidrule(lr){4-5} \cmidrule(lr){6-7}

  & \multirow{2}{*}{\textbf{LL}} & \textbf{Head Ori.} & \multirow{2}{*}{\textbf{LL}} & \textbf{Head Ori.} & \multirow{2}{*}{\textbf{LL}} & \textbf{Head Ori.} \\
  & & MAE (\degree) & & MAE (\degree) & & MAE (\degree) \\

\midrule
  NP & 0.28~(0.24) & 19.63~(7.26) & 0.51~(0.21) & 19.15~(7.22) & 0.54~(0.21) & 18.12~(6.93) \\
  
\specialrule{\lightrulewidth}{1ex}{1ex}
  SP (MLP)  & 0.36~(0.20) & 19.46~(7.05) & 0.52~(0.21) & 19.07~(6.99) & 0.53~(0.21) & 18.65~(6.95) \\
  SP (GRU)  & \textbf{0.55~(0.23)} & \textbf{18.55~(7.11)} & \textbf{1.37~(0.01)} & \textbf{1.11~(0.49)} & \textbf{1.37~(0.01)} & \textbf{0.90~(0.37)} \\
\bottomrule
\end{tabular}
}

%% file: appendix.tex
\onecolumn

\begin{center}
\Large
\textbf{Social Processes: Probabilistic Meta-learning for Adaptive Multiparty Interaction Forecasting} \\
\smallskip Appendices \\
\smallskip
\end{center}

\setcounter{page}{1}

\section{Detailed Results} \label{app:results}

\subsection{Forecasting Glancing Behavior: Quantitative Results} \label{app:toy-quant}
\begin{figure*}[!htb]
  \centering
  \includegraphics[width=0.95\textwidth]{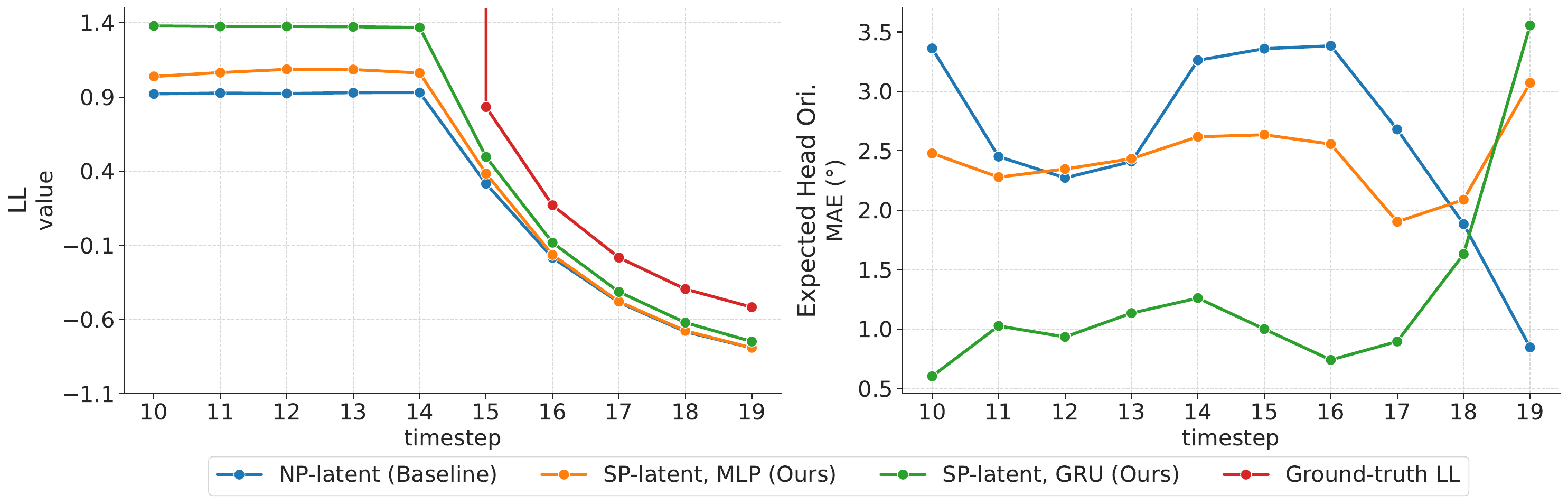}
  \caption{\textbf{Mean Per Timestep Metrics over the Sequences in the Synthetic Glancing Dataset.} We repeat \autoref{fig:ts-synthetic-ll} here for completeness. Head orientation error is computed between the predicted and expected mean (mean of the two ground-truth futures). We observe that the SP-GRU model performs best, especially when the future is certain, learning both the best mean and std. over those timesteps.}
  \label{fig:ts-synthetic}
\end{figure*}

\begin{figure}[!htb]
  \centering
  \includegraphics[width=\textwidth]{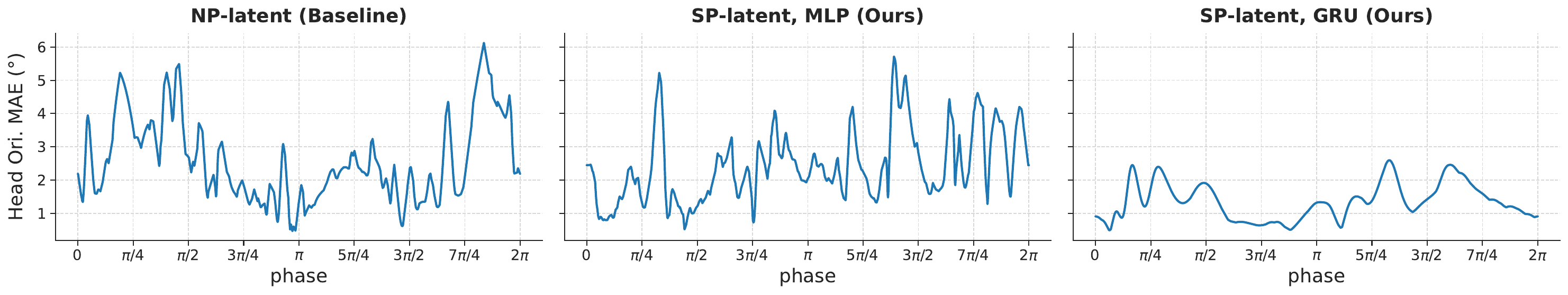}
  \caption{Error in forecast mean and expected mean orientation (mean of the two ground-truth futures) averaged over $\tfut$ for every sequence in the Synthetic Glancing dataset. Each sequence is denoted by the phase of the sinusoid. The SP-GRU error plot is smoother with respect to small phase changes, with lower errors overall.}
  \label{fig:glancing-phases}
\end{figure}

All models are evaluated under the \textit{random} context regime and \textit{no-pool} configuration. The sinusoids are interpreted to represent a horizontal head rotation between $-90\degree$ and $90\degree$. \autoref{fig:ts-synthetic} plots the LL and head orientation error per timestep in $\tfut$. In \autoref{fig:glancing-phases} we plot the MAE in predicted and expected mean forecasts.

\section{Additional Dataset Details} \label{app:data}

\subsection{Synthesized Glancing Behavior Dataset} \label{app:toy-set}
The set of pristine sinusoids representing \textit{Type I} glances is computed by evaluating the sine function at the bounds of $19$ equally spaced partitions of $[0, 3\pi + \phi)$, for phase values $\phi$ in $[0, 2\pi)$ with a step size of $0.001$. More concretely, this is the set
\begin{multline}
    g = \{ r: r = \sin(x),\ x = n \times (3\pi + \phi) / 19,\ n \in \{0, 1, \ldots 19\},\
    \phi = p \times 0.001,\ p \in \{0, 1, \dots 6283\}\},
\end{multline}
which results in $6284$ sequences. \textit{Type III} glances are represented by identical sinusoids with clipped amplitudes for the last six timesteps, resulting in the final dataset of $12568$ sequences. We train with batches of $100$ sequences, using a randomly sampled $25~\%$ of the batch as context. For evaluation, we fix $785$ randomly sampled phase values as context. For each phase, samples corresponding to both types of glances are included in the context set, effectively using $25~\%$ of all samples as context at evaluation.